%% file: main.tex
\documentclass[10pt,twocolumn,letterpaper]{article}
\usepackage[pagenumbers]{cvpr} 
\input{preamble}
\definecolor{customblue}{rgb}{0.21,0.49,0.74}
\usepackage[pagebackref,breaklinks,colorlinks,allcolors=customblue]{hyperref}

\title{BLADE: Single-view Body Mesh Learning through Accurate Depth Estimation}

\author{Shengze Wang\\
{\small UNC (shengzew@cs.unc.edu)}
\and
Jiefeng Li\\
{\small NVIDIA (jiefengl@nvidia.com)}
\and
Tianye Li\\
{\small NVIDIA (tianyel@nvidia.com)}
\and
Ye Yuan\\
{\small NVIDIA (yey@nvidia.com)}
\and
Henry Fuchs\\
{\small UNC (fuchs@cs.unc.edu)}
\and
Koki Nagano*\\
{\small NVIDIA (knagano@nvidia.com)}
\and
Shalini De Mello*\\
{\small NVIDIA (shalinig@nvidia.com)}
\and
Michael Stengel \\
{\small NVIDIA (mstengel@nvidia.com)}\vspace{-2em}
\and
\small{* equal contribution}
}

\begin{document}

\setlength{\abovedisplayskip}{3pt}
\setlength{\belowdisplayskip}{3pt}

\input{sec/1_title_and_abstract}

\input{sec/2_intro}
\input{sec/3_related}
\input{sec/4_method}
\input{sec/5_exp}
\input{sec/6_conclusion}
{
    \small
    \bibliographystyle{ieeenat_fullname}
    \bibliography{main}
}

\clearpage
\input{sec/supplemental.tex}

\end{document}

%% file: preamble.tex
\usepackage[dvipsnames]{xcolor}
\usepackage{multirow}
\usepackage{array} 
\usepackage{graphicx}
\usepackage{keyval}
\usepackage{subcaption}
\usepackage{soul}
\usepackage{tabularx}
\usepackage{changepage}

\newcommand{\smplshape}[1]{\beta_{#1}}
\newcommand{\smplpose}[1]{\theta_{#1}}
\newcommand{\posenet}{$F^{pose}$}
\newcommand{\depthnet}{$F^{T_z}$}
\newcommand{\ourmethod}{BLADE }
\newcommand{\ourmethodnospace}{BLADE}
\definecolor{bluebox}{RGB}{120, 196, 240} %
\definecolor{graybox}{RGB}{120, 120, 120} %
\definecolor{brownbox}{RGB}{209, 188, 105} %

%% file: sec/1_title_and_abstract.tex
\twocolumn[{%
\renewcommand\twocolumn[1][]{#1}
\vspace{-6em}
\begin{adjustwidth}{-0.3in}{-0.3in}
\maketitle
\end{adjustwidth}
\vspace{-2.0em}
\hspace{-3em}
\includegraphics[width=1.1\textwidth, trim=0 0 0 0, clip]{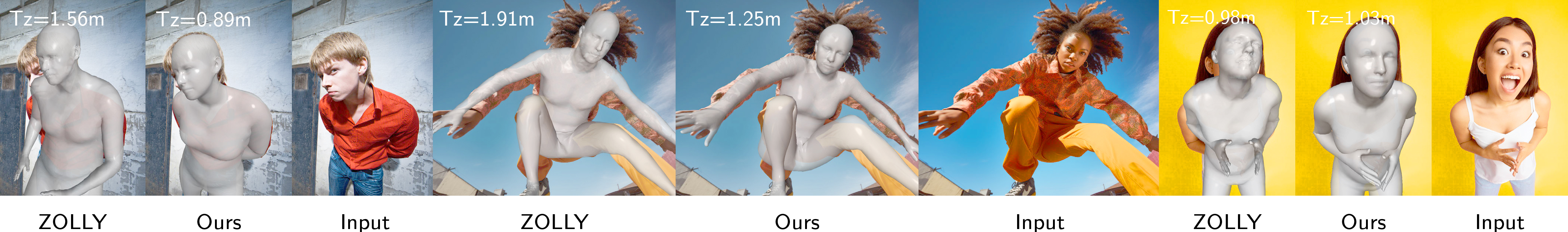} 
\vspace{-1.0em}
\captionof{figure}{
Our method enables accurate human mesh and camera parameter estimation for single-view in-the-wild images including close-ups with high levels of perspective distortion (pelvis depth $T_z$ shown in meters). 
\vspace{2em}
}
\label{fig:teaser}
}]

\begin{abstract}
Single-image human mesh recovery is a challenging task due to the ill-posed nature of simultaneous body shape, pose, and camera estimation.
Existing estimators work well on images taken from afar, but they break down as the person moves close to the camera.
Moreover, current methods fail to achieve both accurate 3D pose and 2D alignment at the same time.
Error is mainly introduced by inaccurate perspective projection heuristically derived from orthographic parameters.
To resolve this long-standing challenge, we present our method BLADE which accurately recovers perspective parameters from a single image without heuristic assumptions.
We start from the inverse relationship between perspective distortion and the person's Z-translation $T_z$, and we show that $T_z$ can be reliably estimated from the image.
We then discuss the important role of $T_z$ for accurate human mesh recovery estimated from close-range images.
Finally, we show that, once $T_z$ and the 3D human mesh are estimated, one can accurately recover the focal length and full 3D translation.
Extensive experiments on standard benchmarks and real-world close-range images show that our method is the first to accurately recover projection parameters from a single image, and consequently attain state-of-the-art accuracy on 3D pose estimation and 2D alignment for a wide range of images.
\url{https://research.nvidia.com/labs/amri/projects/blade/}
\end{abstract}
\vspace{-1em}

%% file: sec/2_intro.tex
\section{Introduction}
\label{sec:intro}

Recent advances in 3D human mesh recovery (HMR) have started to democratize motion capture for media production, allowed computers to understand human gestures for human-computer interaction and enabled new applications in healthcare, fitness, and virtual try-on for E-commerce.
Despite the many successes, current methods struggle in scenarios such as video conferencing and large-scale pose estimation on diverse images captured in the wild (Fig.~\ref{fig:teaser}).

Single-image human mesh recovery is challenging due to the under-constrained nature of estimating many parameters from a single view.
Scale ambiguity and the unknown shape of the person contribute to the existence of potentially an infinite number of valid yet incorrect solutions~\cite{dwivedi_cvpr2024_tokenhmr}.
Furthermore, intrinsic and extrinsic camera parameters are unknown for in-the-wild images and need to be estimated in addition to human shape and pose.
It is thus exceptionally difficult to jointly estimate all of these variables at once.

Therefore, most existing methods reduce the number of unknowns by assuming near-orthographic projection, where the person is assumed to be far away and focal length is heuristically determined or calculated~\cite{li2022cliff,kanazawaHMR18,kocabas2021spec,wang2023zolly,kolotouros2019learning,kocabas2021pare,li2020hybrik}. 
This leads to an unsatisfactory result, especially for close-ups that show a person with strong perspective distortion (Fig.~\ref{fig:teaser}).
Recent work SPEC~\cite{kocabas2021spec} targets this problem by directly estimating the camera focal length from images.
ZOLLY~\cite{wang2023zolly} estimates both the depth of the person and a 2D affine transformation for an orthographic camera, which are then heuristically converted to a focal length and 3D translation with perspective projection.
Both methods rely on inaccurate assumptions and fail to accurately recover the perspective parameters. 

\begin{figure*}[t!]
    \centering
    \includegraphics[width=\textwidth, trim=0 310 0 0, clip]{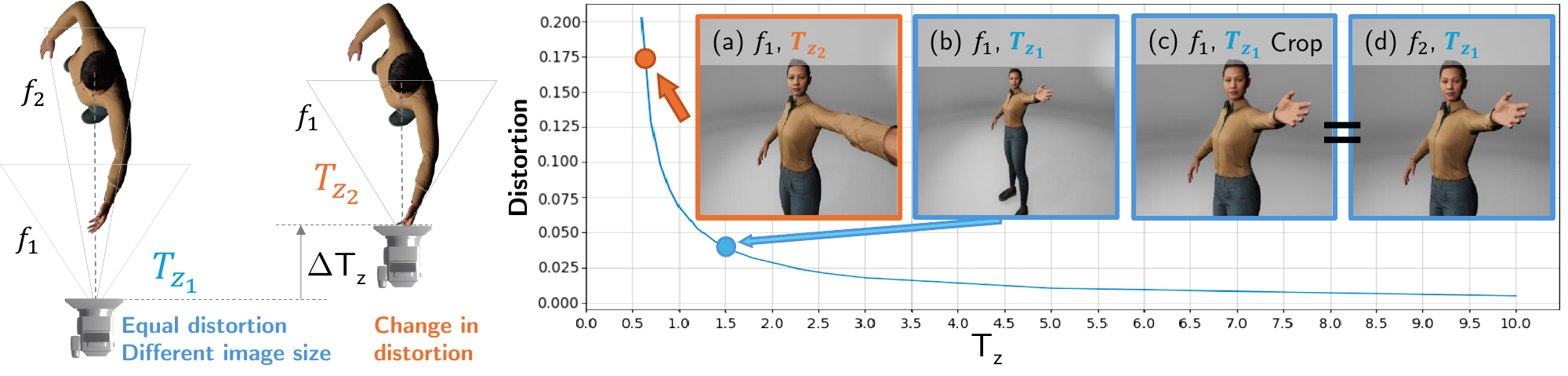}
    \vspace{-1.8em}
    \caption{
    \textbf{Influence of~~$T_z$ on perspective distortion.} A person is captured with different focal length and Z-translation $T_z$ from the camera. (b\&d) Changing the focal length from a short lens $f_1$ to a long lens $f_2$ changes the zoom factor but does not change the perspective distortion, as shown by the equivalence between (c) and (d). (a) Changing the Z-translation by a $\Delta T_z$ changes the level of perspective distortion in the image. This effect is particularly pronounced for close-range imagery (\textcolor{NavyBlue}{blue} curve). See Sec.~\ref{sec:preliminary} for detailed discussion.
    }
    \label{fig:tz_perspectivedistortion}
    \vspace{-1.2em}
\end{figure*}

To simultaneously solve these manyfold challenges, we propose a new method for Body mesh Learning through Accurate Depth Estimation from a single image (\ourmethodnospace).
Our key observation is that, mathematically, perspective distortion is driven by the distance between camera and person, but not affected by focal length (Fig.~\ref{fig:tz_perspectivedistortion}).
The idea is that the Z-translation $T_z$ of the person can be disentangled from other variables and be reliably estimated from the input image (Sec.~\ref{sec:depthestimator}).
Once $T_z$ is estimated, other variables become easier to solve.
Motivated by this intuition as well as the success of recent one-shot metrical depth estimators~\cite{depth_anything_v2, oquab2023dinov2, bochkovskii2024depth}, we train a $T_z$ estimator to predict the depth of the person's pelvis with respect to the camera.
We notice that that human pose estimators predict 3D human mesh from images that are affected by perspective distortion and that perspective distortion is determined by $T_z$.
Therefore, we condition our pose estimator on $T_z$ in order to improve accuracy of estimated human mesh.
Lastly, the focal length and remaining translation parameters $T_x$ and $T_y$ can be obtained with knowledge of $T_z$ and the 3D human mesh shape.
Existing labeled datasets for HMR lack close-range images with strong perspective distortion. To augment them, we also contribute a new large-scale synthetic dataset with 2 million images tailored to this task. It helps our model learn accurate Z-translation of the human body and 3D pose across a wide range of depths.

\begin{figure}[t]
    \includegraphics[width=0.48\textwidth, trim=0 220 350 0, clip]{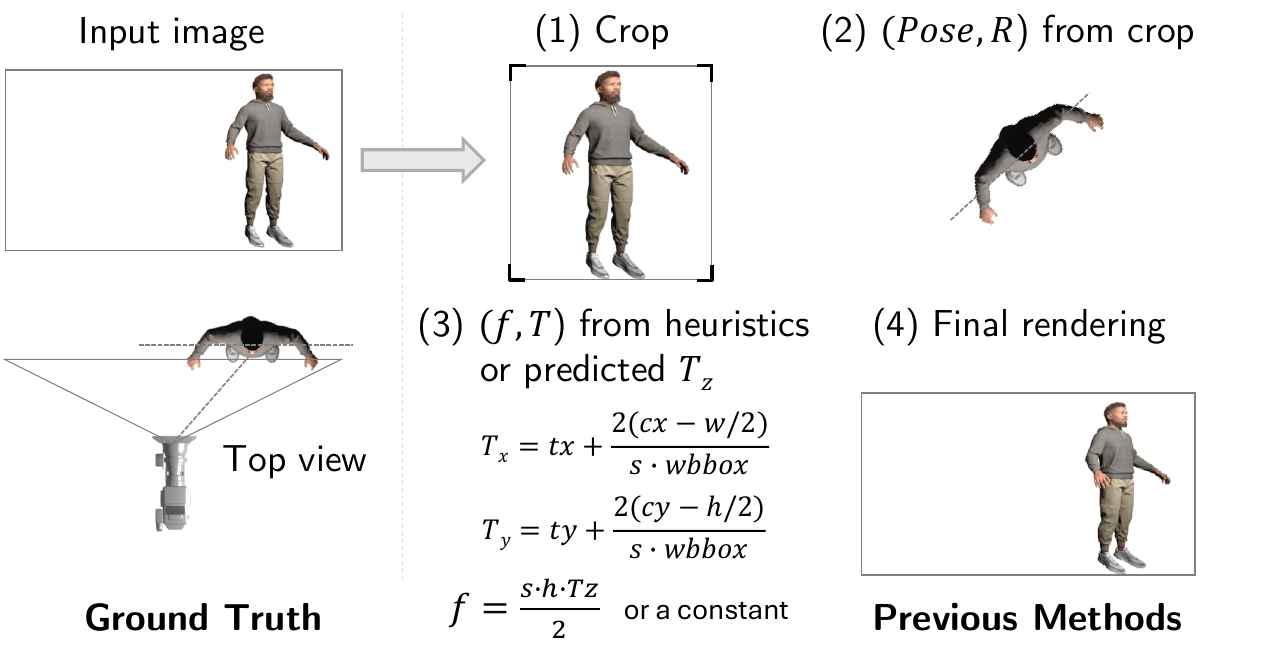}
    \vspace{-1.5em}
    \caption{\textbf{Pose error introduced by camera heuristics.} (1,2) Previous methods estimate the pose of the person from image crops, leading to pose inaccuracy compared to the ground truth (left). (3) Focal length and 3D translation $(f,T)$ are heuristically converted from a 2D affine transformation $(s,t_x,t_y)$, which is only suitable from afar but not for close-range images. (4) Due to the incorrect pose and perspective parameters, the final estimation is inaccurate.}
    \label{fig:persp_vs_ortho}
    \vspace{-1em}
\end{figure}

On several benchmark datasets captured at diverse ranges, we outperform all existing SOTA methods at estimating subject depth, focal parameters, 3D pose, and 2D alignment.
Our work contributes a new angle on accurate single-image 3D human pose estimation.
It is the first method to fully depart from the orthographic camera model and recover a fully perspective projection model without heuristics (Fig.~\ref{fig:persp_vs_ortho}), achieving high accuracy on 3D pose and 2D alignment on diverse depth ranges, including close-range images (Fig.~\ref{fig:teaser} and \ref{fig:results}). 

In summary, we contribute:
\begin{enumerate}
    \item A method for HMR that directly estimates perspective projection parameters given a single image without relying on heuristics. Our method achieves SOTA results on diverse depth ranges, including close-range images.
    \item We identify that close-range pose estimation is heavily affected by Z-translation $T_z$, and we propose to condition the pose estimation on the estimated $T_z$ to improve the accuracy of mesh recovery.
    \item We correct the misconception that focal length affects image distortion, and we show the benefit of estimating focal length and XY-translation independently from $T_z$ and mesh shape and pose.
    \item We contribute a new large-scale synthetic dataset with a wide $T_z$ variety.
\end{enumerate}

%% file: sec/3_related.tex
\section{Related Work}
\label{sec:related}
Human mesh recovery (HMR) from images and video is a long-standing problem and has received broad attention in research.
Tian et al.~\cite{tian2023recovering} provides a comprehensive review of the SOTA in HMR from monocular images. Additional surveys include recovery from multi-view images, videos, and body-worn sensors~\cite{zheng2023deep, dubey2023comprehensive, zhu2023human, liu2024deep}.
In the following, we focus on methods for single-view single-person 3D HMR. This is an important distinction as we target general pose labeling of in-the-wild and internet-scale image datasets for which usually no data beside the images is available.
To obtain realistic and manipulable human bodies, the parametric body model SMPL~\cite{SMPL:2015} and its successor SMPL-X~\cite{SMPL-X:2019} have been proposed.
These models use linear blend skinning for the person's shape along with 3D joint positions and rotations for the pose.

Various methods estimate the body mesh directly using different neural network architectures such as a graph neural network~\cite{kolotouros2019convolutional}, transformer~\cite{cho2022cross}, and a hybrid of the two~\cite{lin2021mesh}.
Other methods regress on the SMPL(-X) body model parameters~\cite{kanazawaHMR18, kolotouros2019learning, zhang2021pymaf, li2022cliff, kocabas2021pare, dwivedi_cvpr2024_tokenhmr, wang23refit} using a multi-stage process that includes cropping of the body parts using detected bounding boxes followed by utilizing distinct models for individual reconstruction of those parts.
In contrast, SMPLer-X~\cite{cai2023smplerx}, OSX~\cite{lin2023osx}, and AiOS~\cite{sun2024aios} regress the body model as a whole, which reduces artifacts stemming from individual part reconstruction.
Additionally, AiOS~\cite{sun2024aios} utilizes a one-stage framework that directly recovers the human mesh from the entire image, omitting body cropping.

Due to the lack of camera information for in-the-wild images, all mentioned methods use orthographic camera models assuming that the person is sufficiently far from the camera. This is not always true in practice.
As shown in Fig.~\ref{fig:persp_vs_ortho}, the weak-perspective assumption often involves estimating a 2D affine transform and heuristically converting the 2D scale and image space translations to focal length and 3D translations. 

Different from these, few prior works do consider perspective distortion~\cite{kissos2020beyond, kocabas2021spec, li2022cliff, wang2023zolly}.
Nagano \etal evaluate the distortion of faces for perspective projection and propose a generative adversarial network to normalize face images with distortion into near-orthographic ones~\cite{nagano2019deep}.
Zhao \etal propose an approach to learning perspective undistortion for face portraits~\cite{zhao2019learning}.
BeyondWeak~\cite{kissos2020beyond} and CLIFF~\cite{li2022cliff} show for HMR that a correction of camera translation from the box crop around the person to the full image improves performance.
BeyondWeak~\cite{kissos2020beyond} also proposes to use a focal length derived heuristically from image resolution as an approximation for the camera field of view (FOV).
SPEC~\cite{kocabas2021spec} predicts camera parameters by learning field of view, camera pitch, and roll.
However, the mentioned methods tend to overestimate focal length and translation and are therefore not reliable for close-up images.

TokenHMR specifically studies the influence of near-orthographic assumptions on the HMR quality~\cite{dwivedi_cvpr2024_tokenhmr}.
TokenHMR reveals that current focal length estimations are inaccurate and unreliable and as a result, improving alignment to the 2D image deteriorates the accuracy of the 3D pose.
It proposes a Threshold-Adaptive Loss Scaling function to achieve both high 2D and 3D accuracy but only for a distant camera.
Our approach is different from TokenHMR as we do not generate perspective projection parameters from an orthographic camera. Instead, we directly solve for precise intrinsic and extrinsic camera parameters.

ZOLLY~\cite{wang2023zolly} is a perspective-aware SOTA method which allows HMR from close-range images.
The method predicts SMPL body parameters inside a bounding box containing the person and estimates the orthographic projection, which is an affine transformation containing a scaling factor $s$.
ZOLLY follows existing heuristics to estimate the focal length as $f=s\cdot h\cdot T_z/2$ and 3D translation as a function of 2D translation and bounding box properties (Fig.~\ref{fig:persp_vs_ortho}).
Here, $h$ is the image height, and $T_z$ is the estimated depth of the SMPL pelvis.
However, these heuristics are inaccurate approximations that lead to incorrect projections.
In this work, we also estimate $T_z$ as part of our method, but we avoid relying on heuristics for estimation.
Instead, we disentangle the parameters to achieve better HMR performance and a more accurate recovery of camera parameters.
There exists no method that can estimate the accurate 3D translation $[T_x,T_y,T_z]$ or correct focal length from a single image.
The problem is inherently ill-posed because there are not enough constraints from a single image to solve for all variables.
On the other hand, significant advancement has been made in solving two major sub-problems, \ie depth estimation \cite{depth_anything_v2,oquab2023dinov2,ke2023repurposing,zoedepth2023,bochkovskii2024depth} and 3D pose estimation \cite{lin2023osx,dwivedi_cvpr2024_tokenhmr,cai2023smplerx,wang2023zolly,kanazawaHMR18,li2022cliff}.
Therefore, we leverage these efforts to solve for the remaining variables, namely $[f,T_x,T_y]$.

\begin{figure*}[ht!]
    \centering
    \includegraphics[width=1.0\textwidth, trim=0 250 0 0, clip]{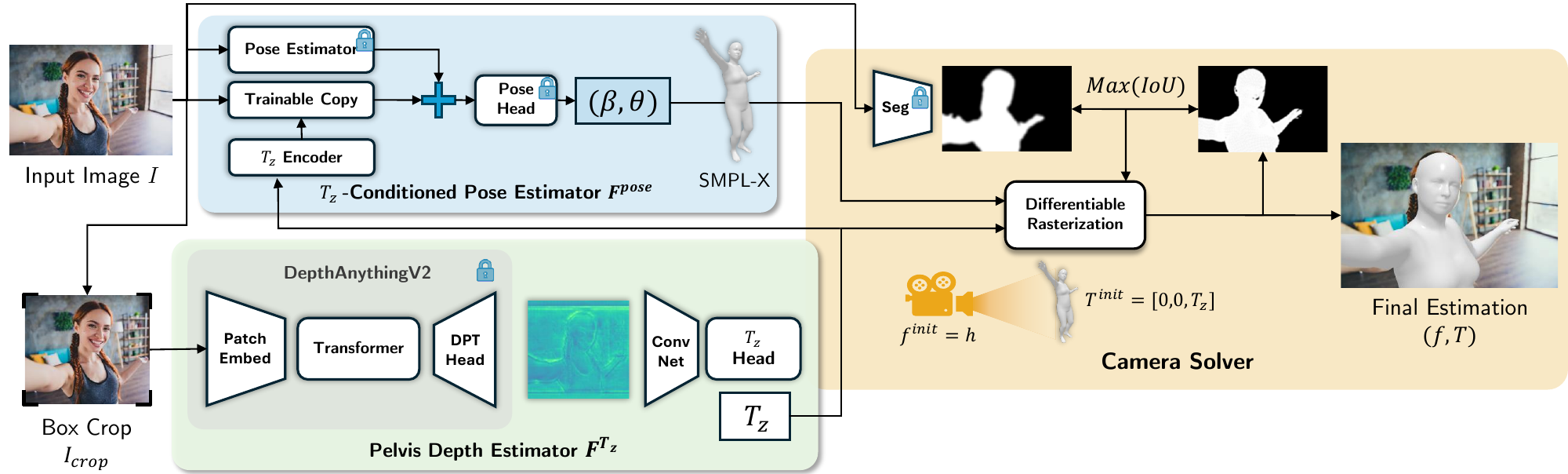}
    \caption{
    \textbf{Overview.} Starting with a bounding box image crop $I_{crop}$ of the person, the \textit{Pelvis Depth Estimator $F^{T_z}$} \textcolor{YellowGreen}{(\textbf{green box})} estimates the Z-translation of the person's pelvis, $T_z$. Then, the \textit{Pose Estimator $F^{pose}$} \textcolor{bluebox}{(blue box)} estimates SMPL-X shape and pose ($\smplshape{}$, $\smplpose{}$) from the full input image while considering the image distortion induced by $T_z$. Finally, through differentiable rasterization, the \textit{Camera Solver} \textcolor{brownbox}{(brown box)} recovers the optimal focal length and 3D translations that best aligns the rasterized SMPL-X mesh with the segmented mask of the person. We are thus able to solve for the full perspective projection model without heuristic assumptions.}
    \label{fig:pipeline}
    \vspace{-1em}
\end{figure*}

%% file: sec/4_method.tex
\section{Method}
\label{sec:method}

Given a single image, our goal is to estimate an accurate 3D mesh of the person as SMPL-X parameters~\cite{SMPL-X:2019} while simultaneously achieving good 2D alignment.
Although it is unreliable to directly estimate camera focal length and extrinsics from a single image, 
we show that they are essentially scaling and alignment parameters, which can be determined once the person's Z-translation $T_z$ is estimated.
Building upon this insight, we introduce a 3-step HMR pipeline (Fig.~\ref{fig:pipeline}) that solves for all essential parameters in perspective projection:
(1) Z-translation $T_z$ of the person with respect to the camera (Sec.~\ref{sec:depthestimator}),
(2) the 3D human pose and shape $(\beta,\theta)$ (Sec.~\ref{sec:poseestimator}),
and finally (3) the person's XY-translations $(T_x, T_y)$ and focal length $f$ (Sec.~\ref{sec:ftxty}).

\subsection{Perspective Projection and its Implication}
\label{sec:preliminary}
SMPL-X provides a differentiable function $M(\beta, \theta)$
that takes the pose parameters $\theta$ and the shape parameters $\beta$ and outputs a body mesh $M \in \mathbb{R}^{N \times 3}$ with $N=10475$ vertices and joint location $J \in \mathbb{R}^{K \times 3}$ with $K=54$ joints.
\footnote{We omit facial expressions and hand gestures due to the lack of such labels in the existing close-range datasets.}
The shape parameters $\beta \in \mathbb{R}^{10}$ are the first 10 PCA coefficients to model body shape variations. 
The pose parameters $\theta \in \mathbb{R}^{3K}$ model the joint rotation including the body orientation.
One can obtain camera space coordinates of SMPL-X vertices $[x_m,y_m,z_m]$ as:
\begin{equation}
    \begin{bmatrix} x, y, z \end{bmatrix}
    = \begin{bmatrix} x_m, y_m, z_m \end{bmatrix}
    + \begin{bmatrix} T_x, T_y, T_z \end{bmatrix},
\label{eq:projection}
\end{equation}
where $T=[T_x,T_y,T_z]$ is the position of the person's pelvis in the camera coordinate. With perspective projection, the projected coordinate is:
\begin{equation}
    \begin{bmatrix}
        u \\
        v 
    \end{bmatrix}
    = 
    f \cdot
    \begin{bmatrix}
        x/z \\
        y/z
    \end{bmatrix}
    =
    f \cdot
    \begin{bmatrix}
        (x_m + T_x)/(z_m + T_z) \\
        (y_m + T_y)/(z_m + T_z)
    \end{bmatrix}.
\label{eq:projection_2}
\end{equation}
According to Eq.~\ref{eq:projection_2}, the projected image coordinate is globally linear with respect to the focal length $f$, indicating that \emph{focal length only acts as a uniform scaling and does not affect perspective distortion.}
In contrast, the distance $T_z$ and 3D geometry, which influence the position $z_m$, have a nonlinear impact on the projected image.
In Fig.~~\ref{fig:tz_perspectivedistortion}, we show how perspective distortion, defined as the difference between perspective and orthographic projection, decreases as $T_z$ increases, whereas perspective distortion quickly increases as $T_z$ decreases in the close range.
This phenomenon presents two key insights:
(1) The amount of perspective distortion observed in an image is strongly correlated to the subject's Z-distance $T_z$ to the camera and hence can be exploited to reliably estimate $T_z$ directly from the image (Sec.~\ref{sec:depthestimator}). 
(2) The same person and pose can result in significantly different projections in the image depending on $T_z$. Thus, when estimating the 3D mesh of the person, the model needs to consider the influence of $T_z$ (Sec.~\ref{sec:poseestimator}).

\subsection{Predicting Z-Translation $T_z$}
\label{sec:depthestimator}
The amount of perspective distortion of a person in an image $I$ is determined by $T_z$, \textit{i.e.}, their distance to the camera (Fig.~\ref{fig:tz_perspectivedistortion}).
Thus, we build a pelvis depth estimator $F^{T_z}$ that directly estimates the depth of their pelvis from their appearance in a cropped image $I_{crop}$ around them, $T_z$ = $F^{T_z}(I_{crop})$.
For $F^{T_z}$ we employ a state-of-the-art pre-trained monocular depth prediction network DAv2~\cite{depth_anything_v2} as a pre-trained backbone to extract appearance features from $I_{crop}$. We find DAv2~\cite{depth_anything_v2} to be the best-performing among several alternatives~\cite{khirodkar2025sapiens, oquab2023dinov2, depth_anything_v2} at this task (Tab.~\ref{tab:ablation_depthbackbone}).
We feed the appearance features into a learnable ConvNet followed by a transformer head module to estimate the pelvis depth $T_z$.
However, as depth can increase to infinity it is impractical to accurately predict depth for the entire unbounded range due to the model's limited learning capacity.
We show in the supplemental material that current backbones struggle to simultaneously achieve high accuracy for both near ranges (\textsc{SPEC-MTP}~\cite{kocabas2021spec}) and farther ranges (\textsc{HuMMan}~\cite{cai2022humman}). 
Hence, it is more important for the model to learn accurate depth prediction for $<$1.2m, where perspective distortion manifests more strongly, versus the farther ranges.
To encourage this, while training $F^{T_z}$ we weigh the $T_z$ error inversely in proportion to the ground truth depth $T_z^{GT}$ resulting in the weighted $L_1$ depth loss:
\begin{equation}
    L_{depth} = 1/T_z^{GT}\cdot \left \Vert T_z - T_z^{GT}\right \Vert_{1}.
\end{equation}


\subsection{$T_z$-aware Pose Estimation}
\label{sec:poseestimator}
As discussed in Sec.~\ref{sec:preliminary} and Fig.~\ref{fig:persp_vs_ortho}, $T_z$ affects the appearance of the human body in the image and thus the accuracy of pose estimation.
Therefore, we design a $T_z$-aware pose estimation block $F^{pose}$ (Fig.~\ref{fig:pipeline}) that takes the input image $I$ and $T_z$ translation to predict the human mesh as SMPL-X parameters, \ie $(\beta, \theta)$.
Specifically, BLADE employs the HMR algorithm AiOS~\cite{sun2024aios}, which directly predicts human meshes from the original uncropped image $I$.
The method extracts features from a pre-trained backbone and contains a transformer-based encoder and non-autoregressive decoder for set prediction of the poses of all persons in an image.
It is trained on large amounts of real-world and synthetic images making it highly generalizable. However, its training data mostly contains distant persons, making it not accustomed to close-range people with strong perspective distortion. 
We find that naively fine-tuning AiOS with smaller close-range datasets employed in~\cite{wang2023zolly} results in over-fitting and undermines its generalizability (Table~\ref{tab:ablation_conditioning}).

To achieve both generalizability and $T_z$-awareness, our pose estimator $F_{pose}$ retains the existing knowledge of the pretrained AiOS while injecting additional depth information $T_z=F^{T_z}(I)$ through a ControlNet~\cite{zhang2023adding} style architecture (Fig.~\ref{fig:pipeline}, pose estimator block).
Specifically, we freeze AiOS and create a trainable copy of its backbone.
The trainable copy is initialized with the pretrained weights, and its output is passed through a zero-initialized MLP before summing with the original output from the frozen backbone.
Before training starts, the zero-MLP creates a zero residual and thus guarantees the same performance as the original AiOS.
Once training starts, the zero-MLP becomes non-zero and allows the trainable backbone to improve upon the original AiOS.
To condition the pose backbone on $T_z$, we use two MLPs to encode $T_z$ into deep features, and we inject the $T_z$ features into the trainable backbone by summing them with the backbone's encoder features.
This way, the existing knowledge is retained in the frozen backbone while the trainable backbone acquires new knowledge about how the $T_z$ distance affects the appearance of the human body in close-range images. 

We input the predicted shape and pose parameters ($\smplshape{}$, $\smplpose{}$) to the SMPL-X function $M$ to obtain the vertices $V$ and joints $J$ with the pelvis joint at the origin:
\begin{equation}
    (\smplshape{}, \smplpose{}) = F^{pose}(I | T_z), \hspace{2em}
    (V, J) = M(\smplshape{}, \smplpose{}).
\end{equation}
To supervise the estimation of human shape, we calculate a shape loss $L_{shape}$ as the $L_1$ distance between the ground truth shape weights $\smplshape{GT}$ and predicted shape parameters $\smplshape{}$:
\begin{equation}
    L_{shape} = L_1(\smplshape{}, \smplshape{GT}).
\end{equation}
To supervise the estimation of pose parameters, we use an angular error between the predicted joint rotations $\smplpose{}$ and ground truth joint rotations $\smplpose{GT}$ (including the root joint orientation):
\begin{equation}
    L_{pose} = E_{ang}(\smplpose{}, \smplpose{GT}).
\end{equation}
We also supervise the position of the estimated SMPL-X joints using a joint location loss $L_{joint}$ as the $L_1$ distance between the predicted joint locations $J$ and ground truth joint locations $J_{GT}$:
\begin{equation}
    L_{joint} = L_1(J, J_{GT}).
\end{equation}
Finally, we supervise the prediction of the mesh vertices by calculating the vertex loss $L_{vert}$ as the distance between ground truth vertices $V_{GT}$ and predicted vertices $V$:
\begin{equation}
    L_{vert} = L_1(V, V_{GT}).
\end{equation}
In summary, the total loss of our pose network is:
\begin{multline}
    L=w_{shape}\cdot L_{shape} + w_{pose}\cdot L_{pose}\\ + w_{joint}\cdot L_{joint} + w_{vert}\cdot L_{vert},
\end{multline}

\noindent where we use $w_{shape}=1$, $w_{pose}=1$, $w_{joint}=5$, $w_{vert}=5$ to balance the magnitudes of the different losses.

\subsection{Solving for Focal Length and 3D Translation}
\label{sec:ftxty}

\begin{figure}[t!]
    \centering
    \includegraphics[width=0.5\textwidth, trim=0 200 0 0, clip]{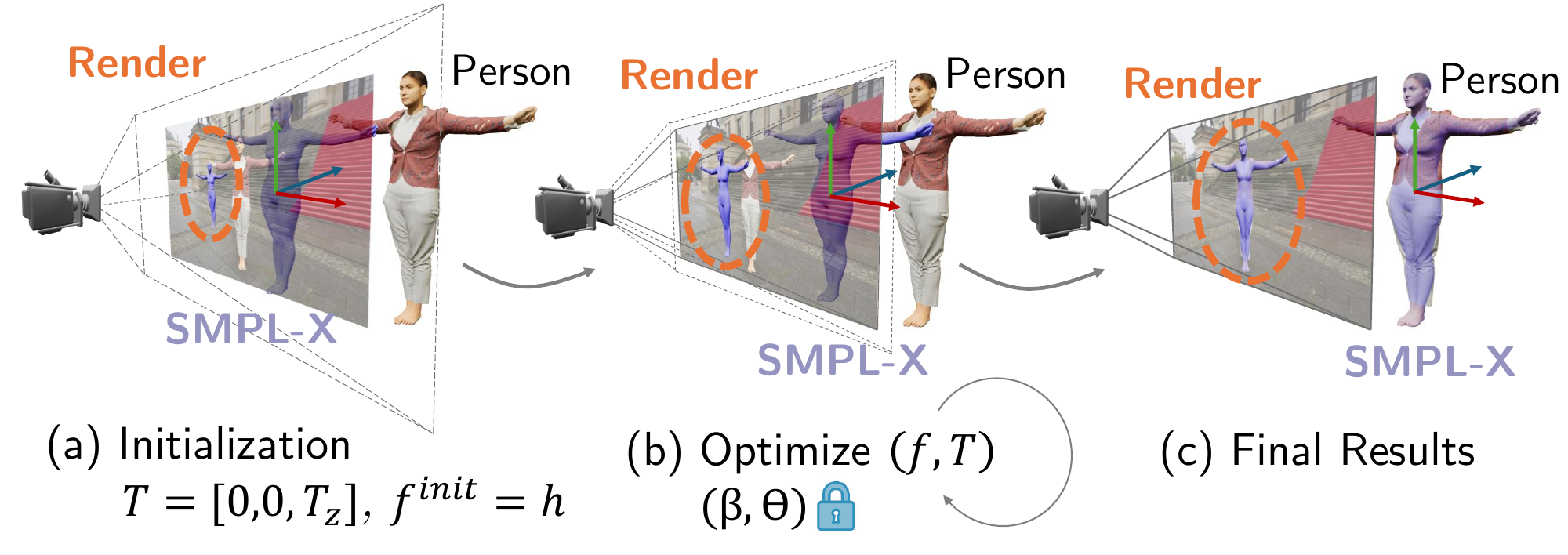}
    \vspace{-1.5em}
    \caption{\textbf{Solving for }$\mathbf{(f,T_x,T_y):}$ (a) With initial $(f,T_x,T_y)=[h,0,0]$, the estimated $T_z$ and human mesh parameters $(\smplshape{}, \smplpose{})$, the optimal $(f,T_x,T_y,T_z)$ is derived (b) by optimizing the image space alignment through differentiable rasterization~\cite{laine2020modular}. (c) The optimized parameters correctly align the projected 3D human mesh to the person in the image.}
    \label{fig:ftxty}
    \vspace{-1em}
\end{figure}

The foundation of our method is the observation that, once $T_z$ is determined, $[f,T_x,T_y]$ can be solved as alignment parameters.
This is because when $T_z$ is fixed, $[T_x,T_y]$ controls movements in the $z=T_z$ plane and $f$ controls the scale of the image.
Therefore, we reformat the problem as an alignment and solve it through differentiable rasterization (Fig.~\ref{fig:pipeline}, brown box).
We render the predicted SMPL-X mesh with an initial translation $T=[0,0,T_z]$ and the initial focal length equals to the image height $f^{init}=h$.
More specifically, we rasterize the SMPL-X model as a binary mask, where pixels are 1 for the projected mesh surface and 0 otherwise.
Then, through differentiable rasterization~\cite{laine2020modular}, we optimize for a tensor $(f,T_x,T_y)$ that maximizes the intersection-over-union between the rasterized SMPL-X mask and the mask of the person, which is segmented using an off-the-shelf method~\cite{lugaresi2019mediapipe}. 
To ensure smooth gradient flow over the entire image, we apply Gaussian smoothing to both the rasterized and segmented masks.
The process is visualized in Fig.~\ref{fig:ftxty} where (1) the purple SMPL-X model shifts to the right such that its projection aligns with the person in the image, and (2) the camera adjusts its focal length to align the sizes of the rasterized and segmented masks.
Additionally, we find that optimizing for $T_z$, and potentially pose and global orientation, often further improves the quality of human pose and camera parameters. 

\subsection{Synthetic Dataset}
\label{sec:bedlamcc}

\begin{figure}[t]
  \centering
\includegraphics[width=\linewidth, trim=0pt 200pt 0pt 40pt, clip]{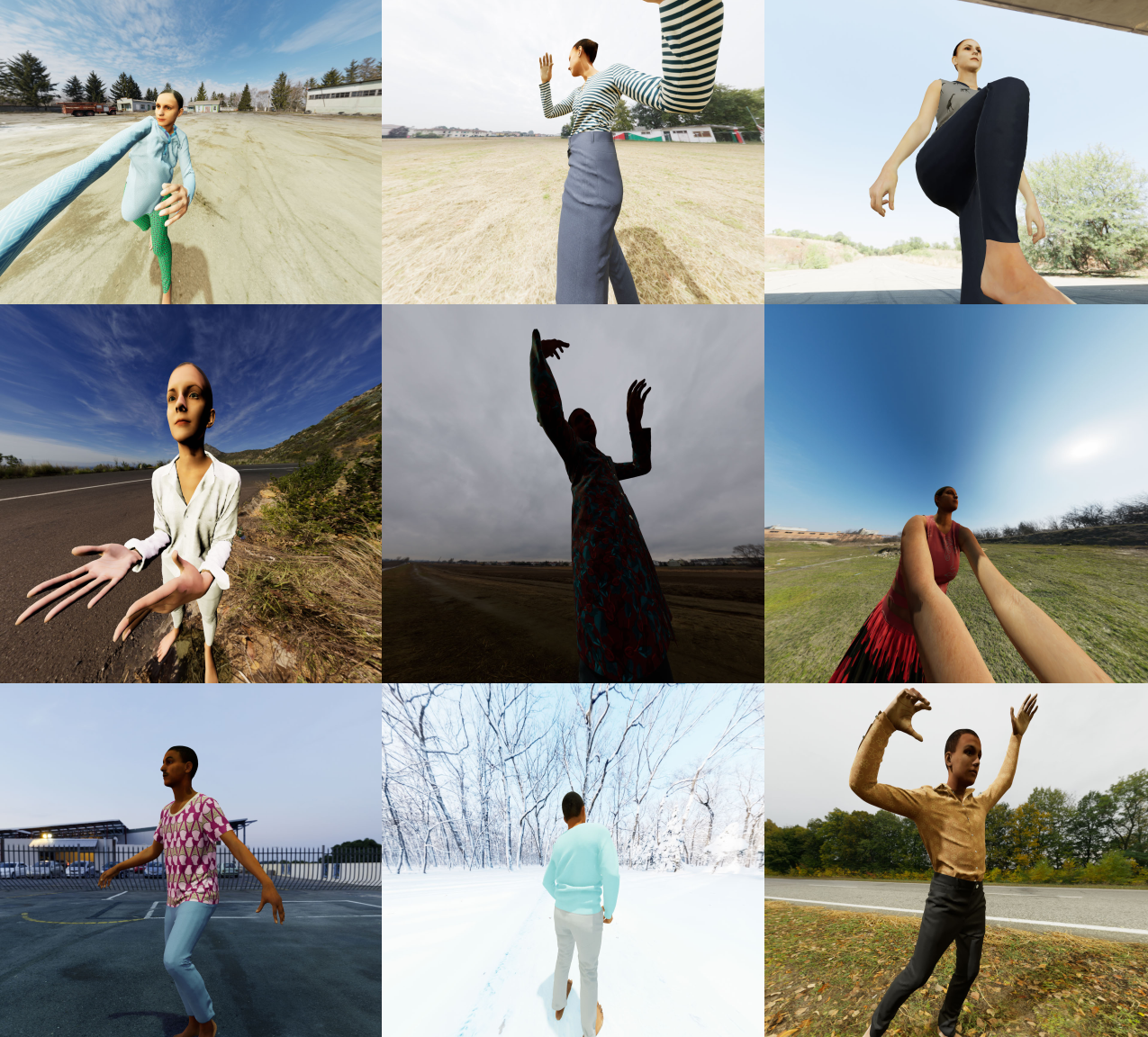}
  \caption{\textbf{Examples of our synthetic \textsc{Bedlam-cc} dataset.} High variation in lighting and camera angles as well as strong close-up distortion are intentionally part of the data.}
  \label{fig:bedlamcc}
  \vspace{-1em}
\end{figure}

While perspective distortion is more severe for the depth range smaller than 1.2m (Sec.~\ref{sec:preliminary}), existing datasets~\cite{cai2022humman,h36m_pami} for HMR do not contain enough data for this range.
An evaluation of $T_z$ distribution for various datasets is included in the supplemental material.
Therefore, we create a new large synthetic dataset we name \textsc{Bedlam-cc} (``close camera'') utilizing assets provided with the \textsc{Bedlam} dataset~\cite{black2023bedlam}.
It contains 2 million synthetically rendered images enhancing current data for depth estimation.
We show example images of our dataset in Fig.~\ref{fig:bedlamcc}.
Focused on challenging close-range images, we uniformly sample the inverse depth $1/T_z$ approximating the perspective distortion curve (Fig.~\ref{fig:tz_perspectivedistortion}) to generate this data.
We enforce that $80\%$ of the samples are within the range of 0.3m $\leq T_z \leq$ 1.2m and the remaining samples in the range of 1.2m $< T_z \leq$ 10m. 
\textsc{Bedlam-cc} is used alongside other datasets to train our Pelvis Depth Estimator $F^{T_z}$.
For fair comparisons during pose estimation, we do not use \textsc{Bedlam-cc} during pose learning.
We also create a separate test set from it for evaluation to provide more accurate ground truth data with a higher depth range.
Please refer to the supplemental material for more details on the \textsc{Bedlam-cc} dataset generation. 

\begin{figure*}[ht]
    \centering
    \includegraphics[width=0.99\textwidth, trim=0 0 0 4.2cm, clip]{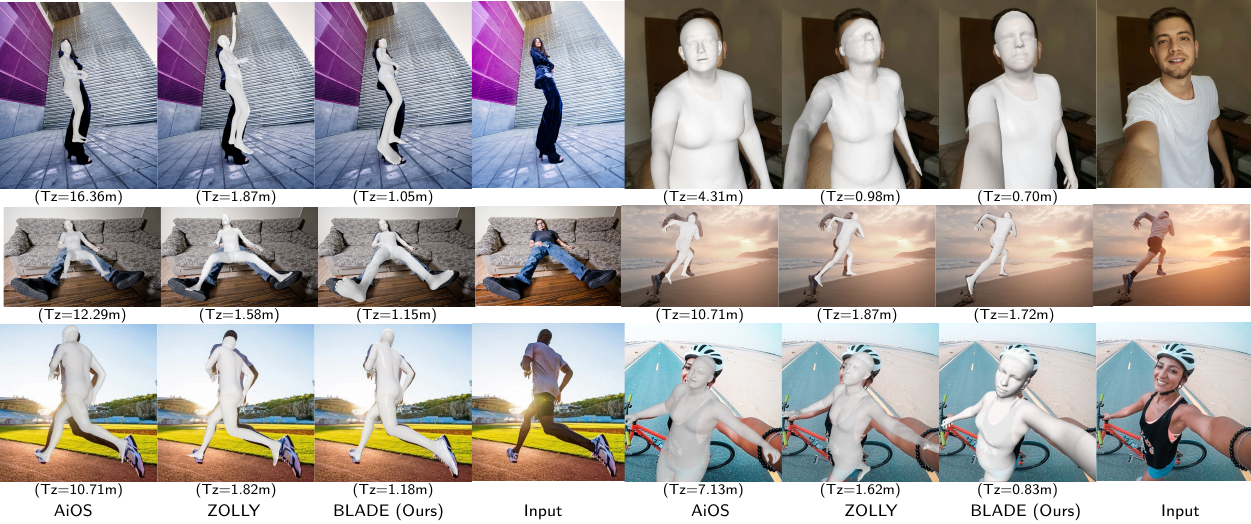}
    \vspace{-0.5em}
    \caption{\textbf{Qualitative SOTA comparison.} We compare with SOTA methods for single-view human mesh recovery including AiOS~\cite{sun2024aios}, and ZOLLY~\cite{wang2023zolly}. 
    Our method \ourmethod is consistently more accurate in terms of estimated pelvis depth $T_z$ of the person (metrical distances given in parenthesis), focal length, and 2D alignment. Notice the improvements for areas with strong perspective effects close to the camera.
    Image sources are given in the supplemental material.
    Images in the bottom row are from ZOLLY~\cite{wang2023zolly}.
    }
    \label{fig:results}
    \vspace{-1em}
\end{figure*}

%% file: sec/5_exp.tex
\section{Experiments}
\label{sec:experiments}
We evaluate our method using existing benchmarks and also present extensive results on real-world images. Our approach recovers both camera parameters and the human mesh, achieving high 3D accuracy as well as precise 2D alignment, whereas prior methods typically excel at only one or the other~\cite{dwivedi_cvpr2024_tokenhmr}.

\subsection{Datasets}
\label{sec:datasets}
We train our model using a subset of 3D datasets employed in ZOLLY~\cite{wang2023zolly}, \ie \textsc{H36M}~\cite{h36m_pami}, \textsc{PDHuman}~\cite{wang2023zolly}, and \textsc{HuMMan}~\cite{cai2022humman}.
These datasets provide labeled camera and SMPL parameters, which we convert to the state-of-the-art SMPL-X model using the method from Choutas \etal~\cite{SMPL-X:2019}.
Following ZOLLY~\cite{wang2023zolly}, we evaluate our method on datasets with strong perspective distortions including \textsc{SPEC-MTP}, \textsc{HuMMan}, \textsc{PDHuman}, and our dataset \textsc{Bedlam-CC}.
\textsc{SPEC-MTP}~\cite{kocabas2021spec} is a real-world dataset with distances ranging from 0.5m to 2m, with most samples captured at approximately 1m.
\textsc{PDHuman}~\cite{wang2023zolly} is a synthetic dataset with distances ranging from 0.5m to 1.8m, where many samples are around 0.6m.
We identified some inconsistencies in the ground truth labels of \textsc{PDHuman}, which we visualize in the supplementary material.
\textsc{HuMMan}~\cite{cai2022humman} is a multi-view dataset captured in a studi, exhibiting limited visual diversity and a narrow distance range of 1.75m to 2.2m. 
To address the above shortcomings, we perform an evaluation on our \textsc{Bedlam-CC} which provides accurate ground truth labels and diverse depth ranging from 0.3m to 10m (Sec.~\ref{sec:bedlamcc}), with $80\%$ of the samples within 1.2m. 
We report performance on HuMMan in the supplementary material, alongside visualizing of depth distributions and inconsistencies in \textsc{PDHuman}.

\subsection{Training}
Our framework contains two modules that require training, namely the pelvis depth estimator \depthnet and the pose estimator \posenet.
We train them in two stages.
During the first stage, we train the pelvis depth estimator \depthnet with a total batch size of 128 on 8 NVIDIA A100 GPUs for 4 epochs.
In the second stage, we freeze \depthnet, feed its prediction of $T_z$ to the pose estimator \posenet, and train \posenet.
The second stage of training uses a batch size of 336 on 48 NVIDIA A100 GPUs for 4 epochs.
The optimization of focal length, and translation vector $T=[T_x,T_y,T_z]$ requires no training.

\subsection{Evaluation Metrics and Baselines}
We evaluate the quantitative performance of all methods using standard metrics and introduce new metrics to evaluate the recovered perspective projection parameters.
We use mean Intersection-over-Union (mIoU) percentage to measure the accuracy of 2D alignment between the rendered mesh and the ground truth mask in the image. 
We use the Per-Vertex Error (PVE) in millimeters to measure the accuracy of the 3D mesh as the $L_2$ distance between the 3D vertices of predicted and ground truth meshes.
We also notice that existing metrics ignore the accuracy of the estimated perspective projection model, which is crucial to achieving consistent 3D pose estimation and 2D pose alignment. 
Therefore, we introduce new metrics to evaluate the accuracy of the recovered perspective projection parameters. 
The common perspective projection model includes focal length and the translation and rotation of the subject in camera space.
We measure the accuracy of the recovered focal length as the percentage error with respect to the ground truth focal length:
\begin{equation}
    E_{f} = |f_{pred} - f_{GT}|/f_{GT}.
\end{equation}
Given that $T_z$ has a direct inverse relationship with the amount of distortion in the image (Fig.~\ref{fig:tz_perspectivedistortion}), whereas $(T_x,T_y)$ do not, we separately evaluate $T_z$ and $(T_x,T_y)$ errors as $E_{T_z}$ and $E_{T_{xy}}$ in meters. 
Additionally, since $T_z$'s accuracy is less important at far distances, we also calculate an inverse $T_z$ error $E_{1/T_z}$ reflecting this property:
\begin{align}
    E_{T_{xy}} &= \|T_{xy}^{pred} - T_{xy}^{GT}\|_2,\\
    E_{T_z} &= |T_z^{pred} - T_z^{GT}|, \\
    E_{1/T_z} &= |1/T_z^{pred} - 1/T_z^{GT}|.
\end{align}
We omit a dedicated 3D rotation error given that 3D rotation is already evaluated as a part of MPJPE.

\begin{table*}[ht]
\centering
{\fontsize{8}{14}\selectfont
\setlength{\tabcolsep}{1pt}
\resizebox{\linewidth}{!}{
\begin{tabular}{lcccccccccccccccccc}
\hline
\textbf{Methods}         & \multicolumn{6}{c}{\textsc{SPEC-MTP}~\cite{kocabas2021spec} (real-world capture)}  & \multicolumn{6}{c}{\textsc{PDHuman}~\cite{wang2023zolly} (synthetic)} & \multicolumn{6}{c}{\textsc{Bedlam-cc} (synthetic)}\\ \cline{2-19}
                & \multicolumn{1}{l}{$E_{T_z}$↓} & \multicolumn{1}{l}{$E_{1/T_z}$↓} & \multicolumn{1}{l}{$E_{T_{xy}}$↓} & \multicolumn{1}{l}{$E_{f}$↓} & \multicolumn{1}{l}{PVE↓} & \multicolumn{1}{l}{mIoU↑} & \multicolumn{1}{l}{$E_{T_z}$↓} & \multicolumn{1}{l}{$E_{1/T_z}$↓} & \multicolumn{1}{l}{$E_{T_{xy}}$↓} & \multicolumn{1}{l}{$E_{f}$↓} & \multicolumn{1}{l}{PVE↓} & \multicolumn{1}{l}{mIoU↑}  & \multicolumn{1}{l}{$E_{T_z}$↓} & \multicolumn{1}{l}{$E_{1/T_z}$↓} & \multicolumn{1}{l}{$E_{T_{xy}}$↓} & \multicolumn{1}{l}{$E_{f}$↓} & \multicolumn{1}{l}{PVE↓} & \multicolumn{1}{l}{mIoU↑} \\ \hline
ZOLLY \cite{wang2023zolly}     & 0.899  & 0.394 & 0.906 & 1.063 & 126.7 & \multicolumn{1}{c|}{62.3}       & 0.255 & 0.355 & 0.267 & 0.273 & 82.0 & \multicolumn{1}{c|}{53.0}    & 0.539 & 0.634 & 0.564 & 0.461 & 131.8 & 51.8 \\
SMPLer-X*\cite{sun2024aios} & 0.980  & 0.450 & 0.109 & 1.121 & 102.6 & \multicolumn{1}{c|}{53.0}       & 2.223  & 1.030 & 0.126 & 0.550 & 161.2 & \multicolumn{1}{c|}{47.6}    & 2.057 & 1.172 & 0.087 & 1.349 & 139.9 & 53.0 \\
TokenHMR*\cite{dwivedi_cvpr2024_tokenhmr} & 0.909 & 0.436  & 0.095 & 1.121 & 124.3 & \multicolumn{1}{c|}{49.7}       & 2.280  & 1.034 & 0.068 & 0.550 & 156.7 & \multicolumn{1}{c|}{53.0}    & 2.378 & 1.200 & 0.096 & 1.349 & 136.4 & 54.2 \\
AiOS*\cite{sun2024aios} & 1.035 & 0.464 & 0.121 & 1.121 & 110.9 & \multicolumn{1}{c|}{48.7}       & 2.312  & 1.024 & 0.149 & 0.550 & 183.4 & \multicolumn{1}{c|}{49.5}    & 2.340 & 1.197 & 0.111 & 1.349 & 143.0 & 54.6 \\
Ours            & 0.129  & 0.114 & 0.056 & 0.163 & 111.9 & \multicolumn{1}{c|}{68.7}       & \textbf{0.106}  & \textbf{0.176} & \textbf{0.043} & \textbf{0.216} & \textbf{80.5} & \multicolumn{1}{c|}{\textbf{67.3}}    & 0.326 & 0.305 & 0.079 & 0.257 & 111.6 & 74.6 \\ 
Ours (real-world)            & \textbf{0.127}  & \textbf{0.112} & \textbf{0.044} & \textbf{0.159} & \textbf{99.6} & \multicolumn{1}{c|}{\textbf{69.5}}       & 0.107  & 0.178 & 0.049 & 0.223 & 102.6 & \multicolumn{1}{c|}{65.2}    & \textbf{0.325} & \textbf{0.305} & \textbf{0.076} & \textbf{0.212} & \textbf{106.8} & \textbf{75.0} \\  
\hline
\end{tabular}
}
}
    \caption{
    \textbf{Quantitative comparison to SOTA methods.} Evaluation on \textsc{SPEC-MTP}~\cite{kocabas2021spec}, \textsc{PDHuman}~\cite{wang2023zolly}, and \textsc{Bedlam-cc}~\cite{black2023bedlam} datasets. Our method achieves SOTA results. Best results indicated by bold numbers. For additional metrics and test datasets please refer to the supplemental material. * symbol indicates pre-trained public models. Model version ``Ours'' is trained using 3D datasets used in ZOLLY~\cite{wang2023zolly} whereas ``Ours (real-world)'' is trained with increased sampling frequency for real-world data \textsc{Human3.6M}~\cite{h36m_pami}.}
    \label{tab:results_selection}
\end{table*}

\subsection{Comparison to State-of-the-Art Methods}
\label{sec:quant_results_sota}

\renewcommand{\arraystretch}{0.9} 
\setlength{\tabcolsep}{3pt} 

\textbf{Quantitative Results}: 
In Table~\ref{tab:results_selection}, we compare our method \ourmethod with state-of-the-art single image HMR methods.
\ourmethod surpasses the current SOTA method for close-range HMR, ZOLLY~\cite{wang2023zolly}, on all datasets and achieves the best overall 2D alignment, 3D localization, and pose estimation.
Notably, \ourmethod obtains a relative improvement of \textbf{85.9}\% $E_{T_z}$ and \textbf{21.4}\% PVE on the \textsc{SPEC-MTP}~\cite{kocabas2021spec} dataset and \textbf{44.8}\% mIoU on the \textsc{Bedlam-cc} dataset.
We also report the performance of recent SOTA methods AiOS~\cite{sun2024aios}, TokenHMR~\cite{dwivedi_cvpr2024_tokenhmr} and SMPLer-X~\cite{cai2023smplerx}, using their respective publicly released models. 
These methods don't explicitly estimate focal length and instead use a constant focal length of 5000.
They estimate accurate 3D meshes with low PVE values but are inaccurate in terms of 2D alignment, focal length and 3D translation.
The common tradeoff between 2D and 3D accuracy is discussed in detail in TokenHMR~\cite{dwivedi_cvpr2024_tokenhmr}.

Additionally, we find that good performance on the synthetic \textsc{PDHuman} dataset~\cite{wang2023zolly} is not representative of good performance in real-world usage.
As shown in Table~\ref{tab:results_selection}, recent SOTA methods~\cite{cai2023smplerx,sun2024aios,dwivedi_cvpr2024_tokenhmr} perform well on the real-world dataset \textsc{SPEC-MTP} but substantially worse on \textsc{PDHuman} in terms of PVE. 
Whereas ZOLLY~\cite{wang2023zolly} performs well on \textsc{PDHuman} but less so on \textsc{SPEC-MTP}~\cite{kocabas2021spec}.
We suspect that this potential domain gap is due to: (1) the extreme distortion in the \textsc{PDHuman} dataset which is not present in real-world data, and (2) inconsistencies in its ground truth labels (detailed in the supplementary).
We thus show two versions of \ourmethodnospace: (i) ``Ours'' trained with a balanced distribution across the 3 training datasets; and (ii) ``Ours (real-world)'' trained with increased sampling from \textsc{Human3.6M} and decreased sampling from \textsc{PDHuman}. ``Ours'' performs well on each dataset compared to other methods and performs best on \textsc{PDHuman}. ``Ours (real-world)'' performs the best on \textsc{SPEC-MTP}, \textsc{Bedlam-cc}, and in real-world usage.
Please refer to the supplementary for an expanded version of Table~\ref{tab:results_selection} with all metrics and additional results.

\noindent\textbf{Qualitative Results}:
In Fig.~\ref{fig:teaser} and Fig.~\ref{fig:results}, we show results of SOTA methods AiOS~\cite{sun2024aios} and Zolly~\cite{wang2023zolly}, and our method on real-world images.
\ourmethod performs significantly better than compared methods in terms of 2D alignment of the mesh to the image, 3D body mesh, and the accuracy of perspective distortion.
The alignment of body parts close to the camera is specifically improved by our method.
More visual results are included in the supplementary.

\subsection{Ablation Study}
\label{sec:ablation}
\paragraph{Ablation of pelvis depth estimator.}
Accurate depth estimation is the core to solving for other variables.
In Table~\ref{tab:ablation_depthbackbone} we evaluate various foundation models including DiNOv2~\cite{oquab2023dinov2}, Sapiens~\cite{khirodkar2025sapiens}, and DAv2~\cite{depth_anything_v2} as the backbone to our pelvis depth estimator \depthnet. 
The models are trained using \textsc{HuMMan}~\cite{cai2022humman}, \textsc{PDHuman}~\cite{wang2023zolly}, and \textsc{Human3.6M}~\cite{h36m_pami}.
On the most challenging real-world \textsc{SPEC-MTP}~\cite{kocabas2021spec} dataset, DAv2 achieves the best accuracy with $E_{T_z}$ = 15.4cm. 
Finally, ``Ours" is a version of the DAv2-based \depthnet{} trained with improved augmentation and additional data from our \textsc{Bedlam-cc} dataset (Sec.~\ref{sec:bedlamcc}), which provides many close-range images ($<$1m), and thus further reduces the $T_z$ error from 15.4cm to 12.7cm.

\renewcommand{\arraystretch}{0.9}
\setlength{\tabcolsep}{3pt}

\begin{table}[t]
    \small
    \centering
    {
    \begin{tabularx}{0.48\textwidth}{p{1.cm}>{\centering\arraybackslash}p{1.8cm}>{\centering\arraybackslash}p{1.5cm}>{\centering\arraybackslash}p{1.9cm}p{1.5cm}}
        \toprule
        \textbf{} &  DiNOv2~\cite{oquab2023dinov2} & Sapiens~\cite{khirodkar2025sapiens} & DAv2~\cite{depth_anything_v2} & Ours \\ \midrule
         \multicolumn{1}{l}{$E_{T_z}\downarrow$} & 0.300 & 0.210 & 0.154 & \textbf{0.127} \\ \bottomrule  
        \end{tabularx}
    \caption{\textbf{Ablation study for depth backbone.} Test on \textsc{SPEC-MTP}~\cite{kocabas2021spec}. ``Ours'' is using DAv2 as the depth backbone~\cite{depth_anything_v2} and fine-tuned using different augmentations.}
    \vspace{-1.5em}
    \label{tab:ablation_depthbackbone}
    }
\end{table}

\vspace{-0.5em}
\paragraph{Conditioning the pose estimator.}
In Table~\ref{tab:ablation_conditioning}, we evaluate various architectures of pose estimator on the task of 3D pose estimation and mesh recovery on the challenging close-range real-world \textsc{SPEC-MTP} dataset~\cite{kocabas2021spec}. The publicly available ``raw AiOS" performs well.
However, after fine-tuning (``ft. AiOS") with the \textsc{HuMMan}, \textsc{PDHuman}, \textsc{H36M} datasets, which mostly contain faraway subjects and synthetic images,
its performance degrades on the close-range real-world \textsc{SPEC-MTP} dataset~\cite{kocabas2021spec}, by losing its good generalization to real-world data.
On the other hand, conditioning raw AiOS~\cite{sun2024aios} in $T_z$ through a ControlNet-style architecture~\cite{zhang2023adding} that we proposed in \ourmethod (Fig.~\ref{fig:pipeline}), leads to significant improvements in pose estimation performance. It enables the pose backbone to retain its previous knowledge while learning the correct relationship between $T_z$ and the image to enhance 3D pose estimation.

\renewcommand{\arraystretch}{0.9}
\setlength{\tabcolsep}{3pt}

\begin{table}[t]
    \small
    \centering
    \begin{tabularx}{0.48\textwidth}{p{2.5cm}>{\centering\arraybackslash}p{2cm}>{\centering\arraybackslash}p{1.7cm}>{\centering\arraybackslash}p{1.2cm}}
        \toprule
        \textbf{}  & PA-MPJPE$\downarrow$ & MPJPE$\downarrow$ & PVE$\downarrow$ \\ \midrule
        raw AiOS   & 62.816 & 101.577 & 110.851 \\
        ft. AiOS & 64.932 & 113.173 & 120.582 \\
        Ours ($T_z$ cond.) & \textbf{56.666} & \textbf{94.050} & \textbf{99.635} \\ \bottomrule        
        \end{tabularx}
    \caption{\textbf{Ablation study for conditioning.} Test on \textsc{SPEC-MTP}~\cite{kocabas2021spec}. Architecture: DAv2~\cite{depth_anything_v2} used in pelvis depth estimator. First row: AiOS~\cite{sun2024aios} used as pose estimator. Second and third row ``Ours'': AiOS~\cite{sun2024aios} with ControlNet~\cite{zhang2023adding} used as pose estimator with and without conditioning on $T_z$.}
    \vspace{-0.5em}
    \label{tab:ablation_conditioning}
\end{table}

%% file: sec/6_conclusion.tex
\paragraph{Limitations.}
We currently only consider single-person images. For the future, we plan to extend our method to process videos where more information can be leveraged for better accuracy. 
We also do not consider lens distortion or camera types other than the standard pin-hole camera such as fish eye lenses. 
Lastly, the estimation of $(f,T_x,T_y)$ can fail when the segmentation mask is very inaccurate. 
A promising direction is learnable optimization to substitute differentiable rasterization for better robustness.

\section{Conclusion}
\label{sec:conclusion}

In this work, we propose \ourmethod -- a method for human mesh recovery and perspective camera estimation from single images. This is a long-standing challenging and open problem.
Different from previous work, we provide a solution to estimating perspective projection parameters without conversion from an orthographic camera model.
We underscore the significance of accurate and disentangled pelvis depth estimation, followed by depth-conditioned human pose estimation, and finally optimization of camera focal length and XY-translation.
We also introduce a large-scale synthetic single-person dataset, \textsc{Bedlam-cc}, containing a large number of close-range images with ground truth labels for the perspective camera and SMPL-X body parameters.
Our framework \ourmethod achieves state-of-the-art accuracy on a variety of benchmarks and across a wide range of depths.
Among other use cases, the method can be applied for accurate pose labeling of in-the-wild image datasets to train robust human-centric models.

%% file: sec/supplemental.tex
\appendix

\renewcommand{\thesection}{A\arabic{section}}
\renewcommand{\thefigure}{A\arabic{figure}}
\renewcommand{\thetable}{A\arabic{table}}

\twocolumn[{%
\renewcommand\twocolumn[1][]{#1}%
\vspace{2em}
\maketitle
\centering
{\Large \bfseries BLADE: Single-view Body Mesh Learning through Accurate Depth Estimation\\\vspace{0.5em}Supplemental Material}\\[1em]
\vspace{1em}
\includegraphics[width=0.99\textwidth, trim=0 120 0 0, clip]{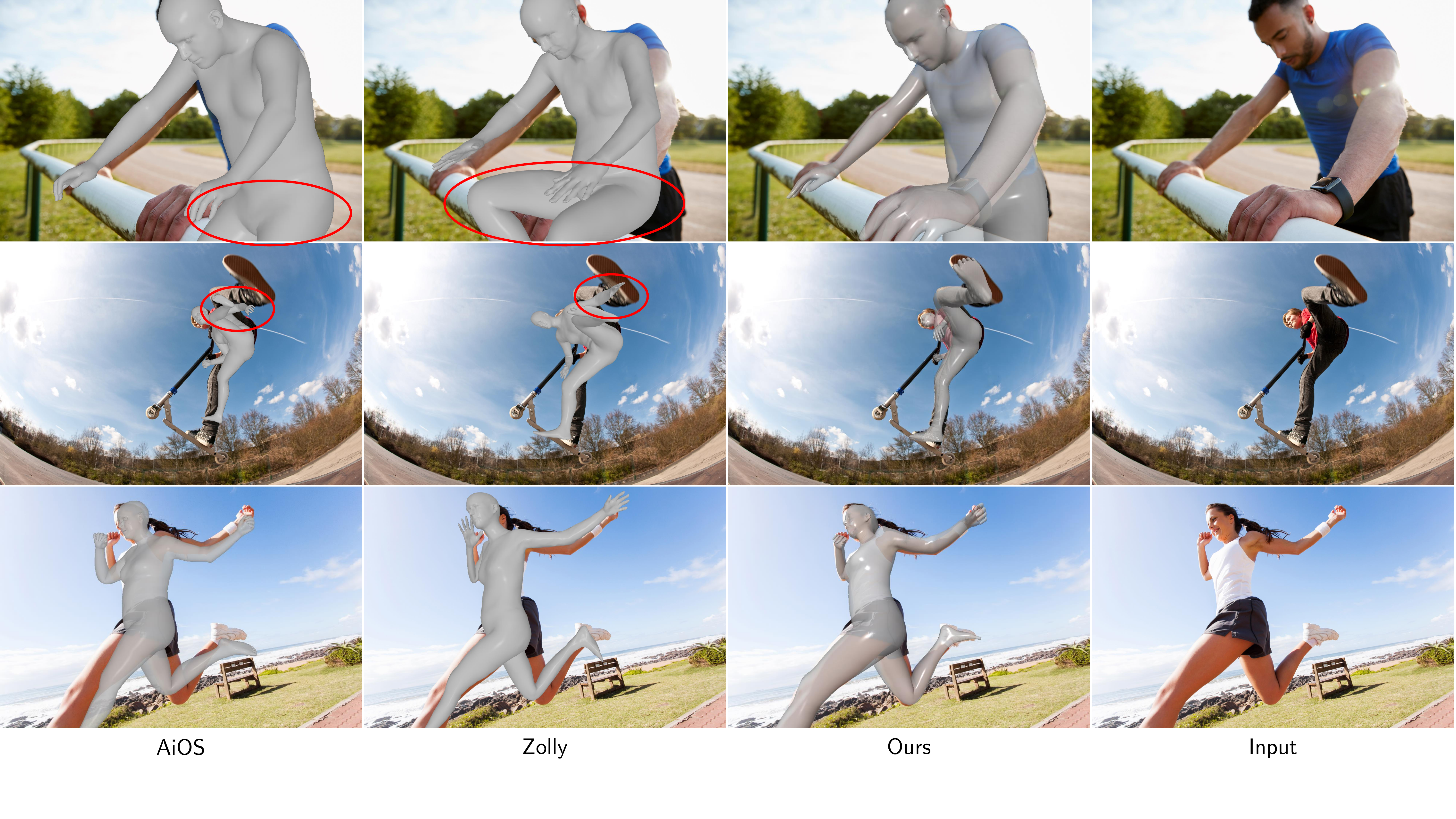} 
\captionof{figure}{\textbf{More Qualitative Results.} BLADE not only achieves accurate 3D pose estimation, but also accurately recovers perspective projection parameters and thus achieves state-of-the-art alignment accuracy in image space.}
\vspace{2em}
\label{fig:vis_1}
}]

\section{Overview}
In this supplemental document, we 
(1) provide additional qualitative results on real-world images (Sec.~\ref{sec:qualitative}); 
(2) examine the existing evaluation datasets and identify the need for a close-range evaluation dataset with accurate labels (Sec.~\ref{sec:datasets}); 
(3) report additional quantitative results of the various methods on more datasets and with additional metrics (Sec.~\ref{sec:results});
(4) elaborate on the ambiguity involved in single-image-based 3D human mesh recovery (Sec.~\ref{sec:ambiguity}); and 
(5) discuss the trade-off between achieving high depth estimation accuracy on close-range data versus far-range data (Sec.~\ref{sec:depthtradeoff}).

\newpage

\section{Qualitative Results on Real-World Images}
\label{sec:qualitative}
In Fig.~\ref{fig:vis_1}, ~\ref{fig:vis_2} and ~\ref{fig:vis_3}, we show more visual results with a comparison to recent state-of-the-art methods AiOS~\cite{sun2024aios} and Zolly~\cite{wang2023zolly}. 
We achieve significant improvement in terms of alignment of the rendered 3D mesh to the input image, accuracy of perspective distortion, as well as the estimated 3D pose.
For example, in the first row of Fig.~\ref{fig:vis_1}, only our method correctly estimates the camera's close proximity to the person's hand and that the person is standing, whereas AiOS and Zolly predict incorrect leg postures and distances to the person.
In the second row of Fig.~\ref{fig:vis_1}, both AiOS and Zolly wrongly estimate the person's left hand behind their body, whereas BLADE recovers the correct position of the person's hand and camera's proximity to the person's feet.
A similar phenomenon can be observed in Fig.~\ref{fig:vis_2}, \ref{fig:vis_3}, \ref{fig:vis_4}, and \ref{fig:vis_5} as well.

Interestingly, Zolly~\cite{wang2023zolly} sometimes generates flattened meshes.
For example, in the second image from top left in Fig.~\ref{fig:vis_2}, Zolly predicts a mesh where the person's head and arms are flattened.
This is because, different from AiOS and our methods, Zolly directly predicts a mesh instead of parameters of the SMPL-X model. 
While this design gives Zolly more flexibility in generating difficult shapes, it can also lead to degenerate estimation at times.

Additionally, although BLADE leverages AiOS~\cite{sun2024aios} as part of the pose estimator backbone, BLADE improves AiOS' pose and shape accuracy.
For example, in the top left of Fig.~\ref{fig:vis_4}, BLADE predicts the person's body shape more accurately than AiOS.
In the second and bottom row in Fig.~\ref{fig:vis_4}, predictions of the person's legs from AiOS and Zolly are both wrong whereas BLADE shows robustness in both situation.
In the top row of Fig.~\ref{fig:vis_5}, BLADE correctly recovers both the orientation and the leg posture of the person, whereas AiOS does not.
In the second row of Fig.~\ref{fig:vis_5}, BLADE correctly recovers the position and angle of the person's ankles, whereas predictions from AiOS are inaccurate.

\section{Examining the Evaluation Datasets}
\label{sec:datasetsextended}
\begin{figure}[t]
    \centering
    \includegraphics[width=0.5\textwidth, trim=0 10 0 0, clip]{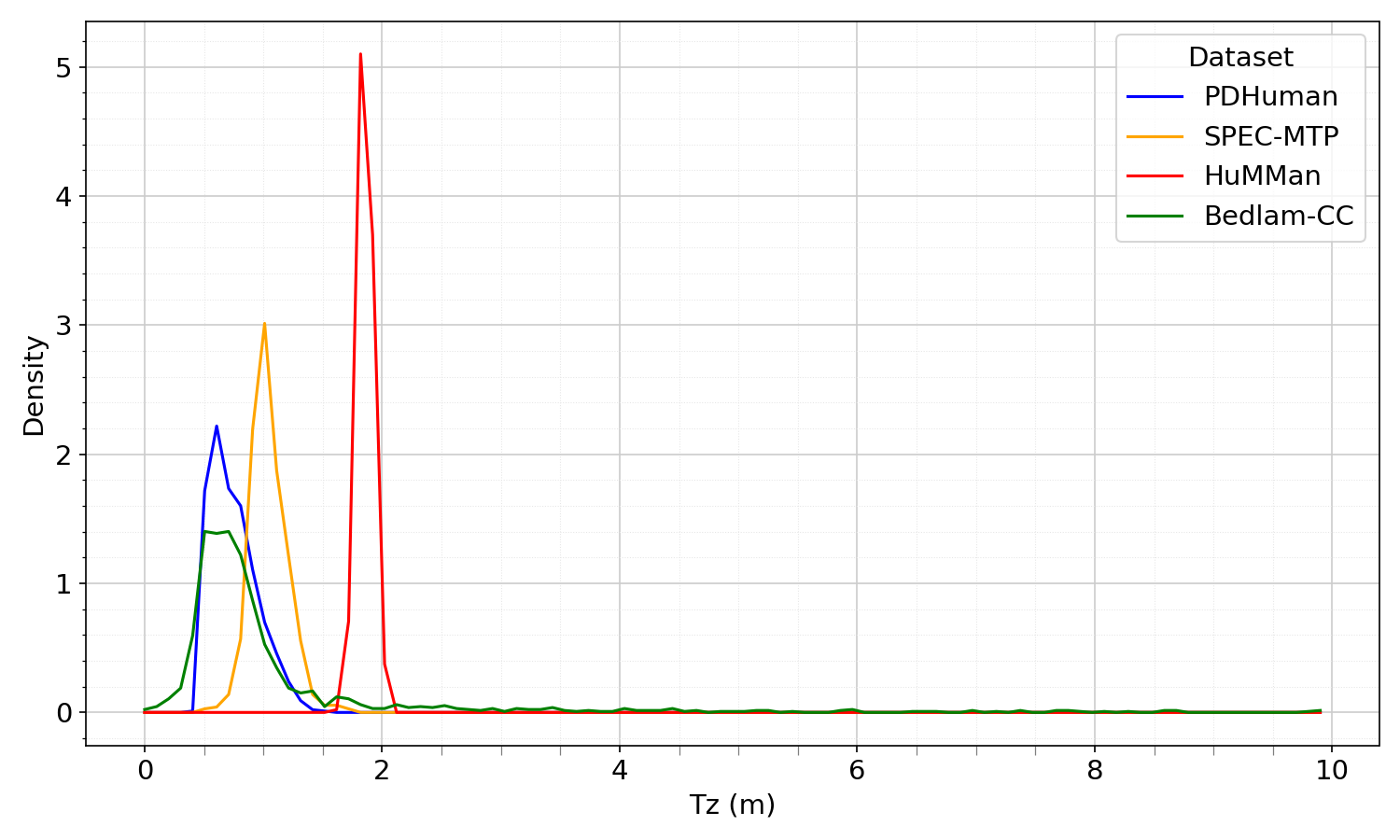}
    \includegraphics[width=0.5\textwidth, trim=0 10 0 0, clip]{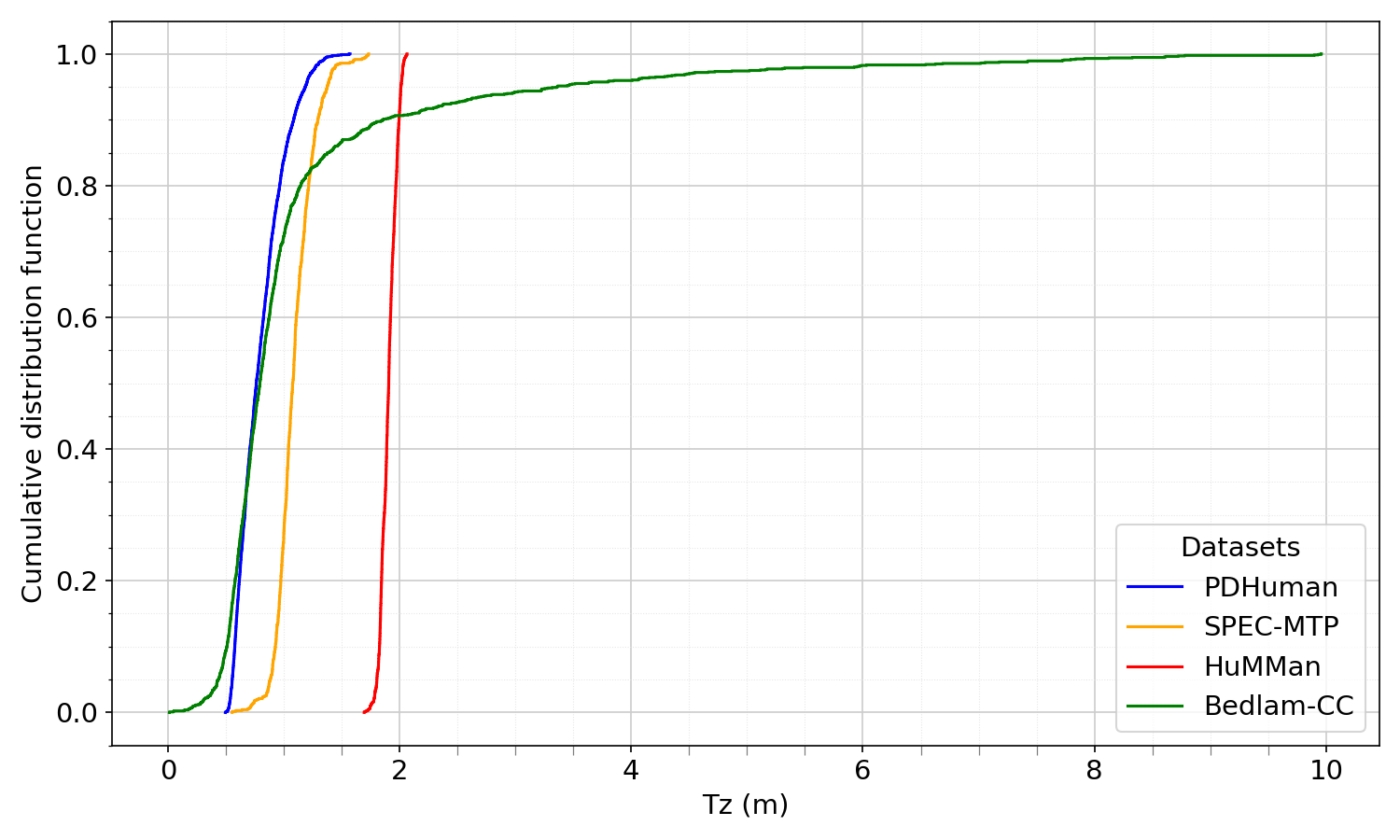}
    \caption{\textbf{Evaluation Dataset Distributions.}
    In the top diagram, we show the distribution of $T_z$ values across different datasets. 
    Notably, the majority of the \textsc{HuMMan} dataset has $T_z$ values concentrated in a small range around 1.9m.
    The \textsc{HuMMan} dataset thus has much less perspective distortion compared to close-range datasets like the \textsc{SPEC-MTP}\cite{kocabas2021spec}, \textsc{PDHuman}\cite{wang2023zolly}, and our \textsc{Bedlam-cc} dataset.
    In the bottom, we show the cumulative distribution function of $T_z$ values across datasets. 
    Notably, our \textsc{Bedlam-cc} dataset has a wider range of $T_z$ values, and even smaller minimum $T_z$ values than \textsc{PDHuman}.
    These traits make \textsc{Bedlam-cc} a diverse evaluation dataset that is particularly well-suited for close-range HMR.
    }
    \label{fig:distribution_histogram}
\end{figure}

\begin{figure}[t]
    \centering
    \includegraphics[width=0.5\textwidth, trim=0 10 0 0, clip]{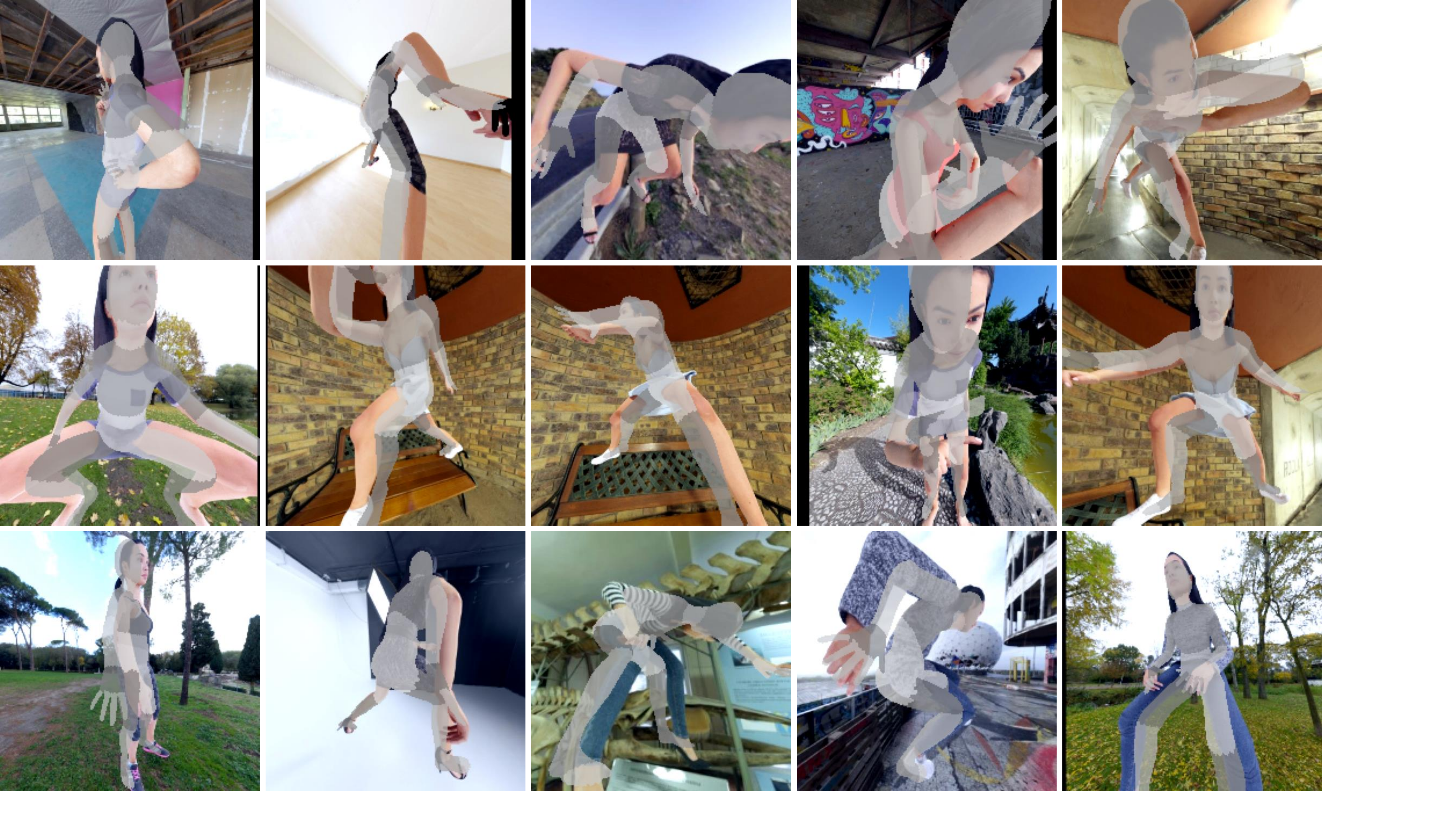}
    \includegraphics[width=0.5\textwidth, trim=0 10 0 0, clip]{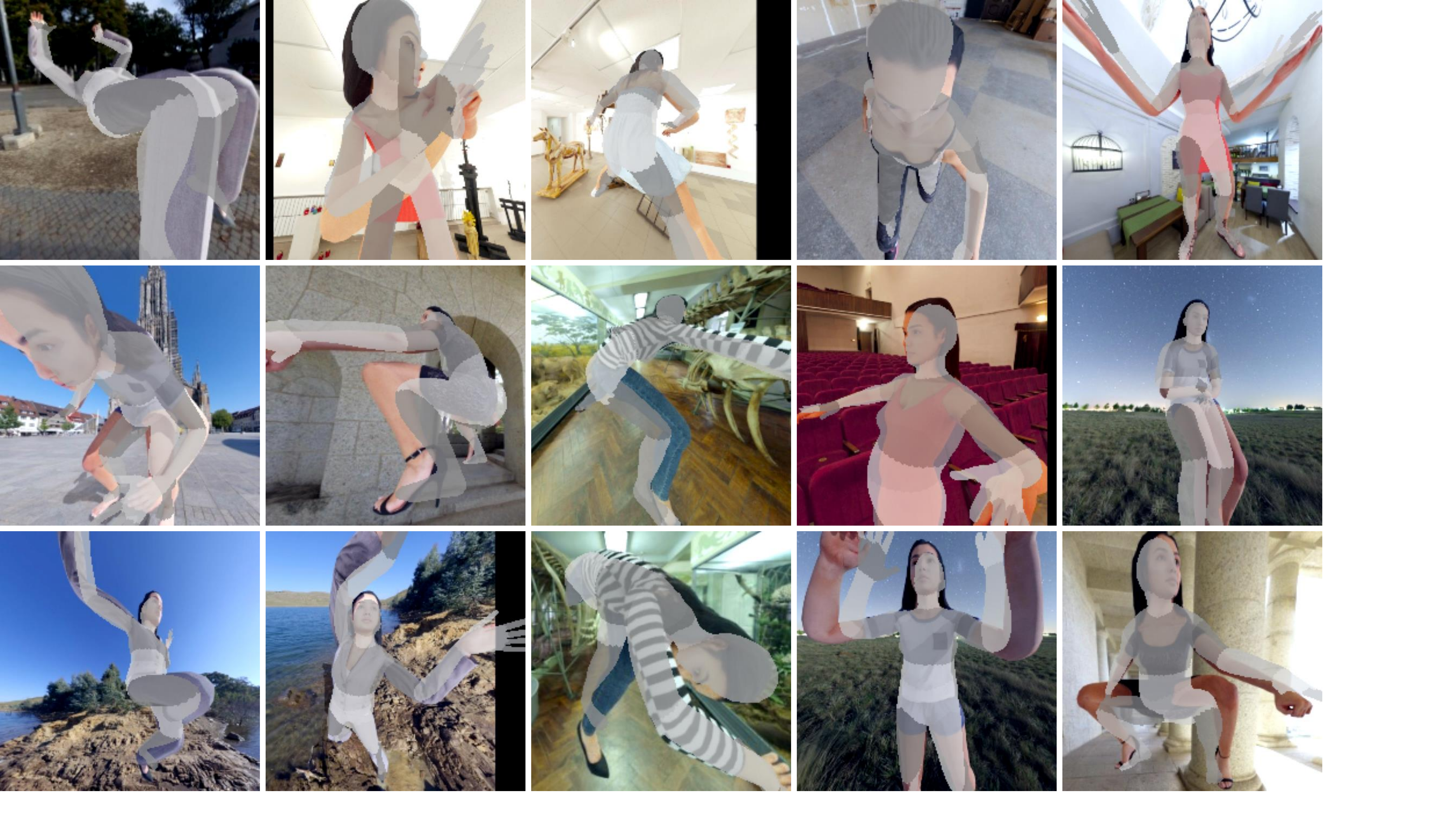}
    \caption{\textbf{Inaccurate Pose Labels in PDHuman\cite{wang2023zolly}.} We find that a high percentage of pose labels in PDHuman do not align with the corresponding images. In the above examples, we visualize the SMPL labels superimposed on top of the corresponding images. The SMPL renderings (gray overlays) are generated using the authors' original code base used for IoU calculations.}
    \label{fig:bad_gt}
\end{figure}

\begin{figure*}[t]
    \centering
    \includegraphics[width=0.8\textwidth, trim=0 120 1400 0, clip]{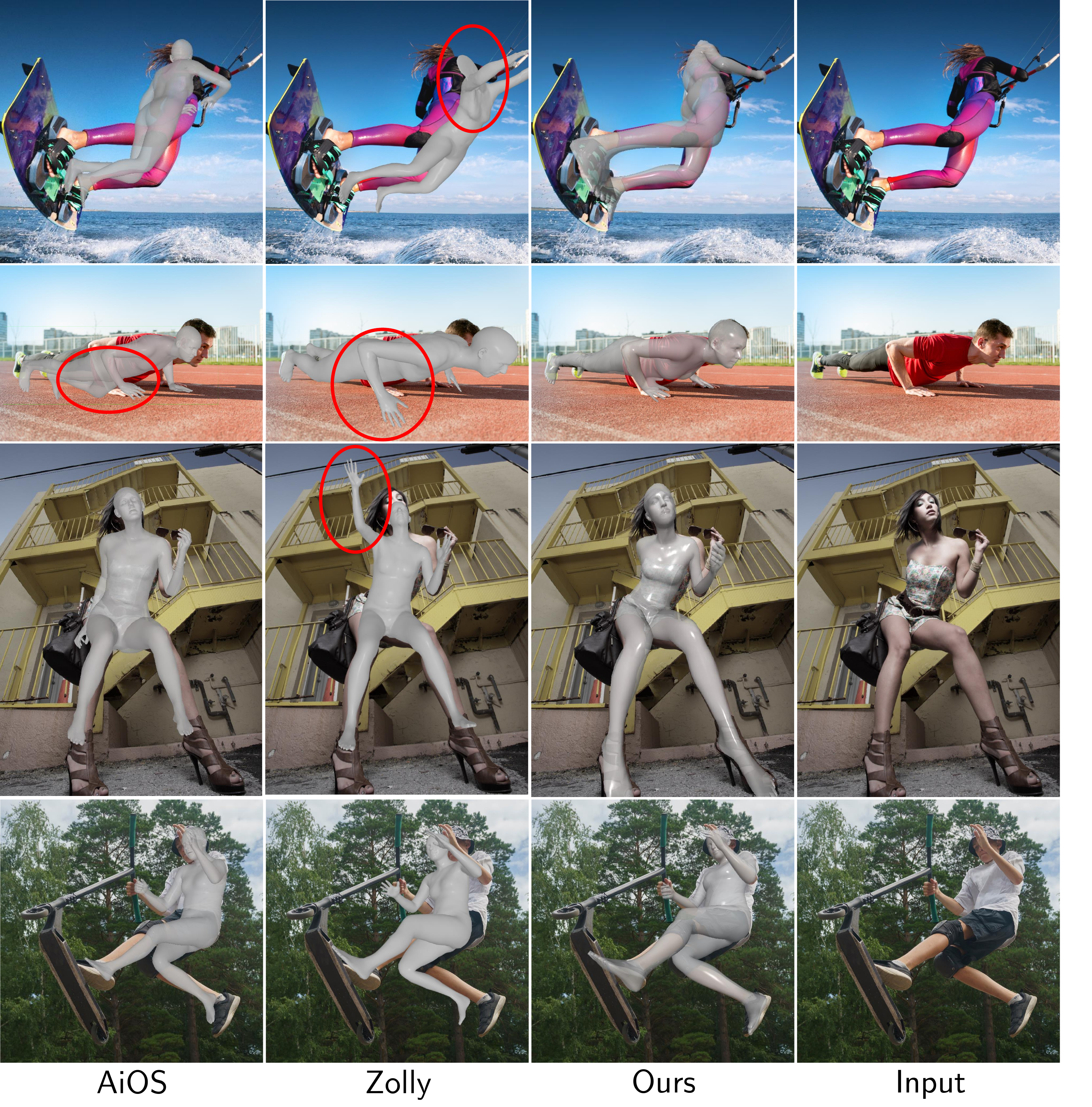}
    \caption{\textbf{More Qualitative Results.} In addition to achieving accurate pose estimation, our method BLADE recovers precise perspective projection parameters, ensuring the predicted 3D human mesh is well-aligned with the input image.}
    \label{fig:vis_2}
\end{figure*}

In this section, we examine the strengths and shortcomings of various standard benchmark datasets used to evaluate the task of single-image-based human mesh recovery (HMR). 
We find that there is a lack of close-range test data with accurate ground truth annotations, and we thus introduce \textsc{Bedlam-cc} to fill this void.

In Fig.~\ref{fig:distribution_histogram}, we show the distribution of $T_z$, \ie the depth of the pelvis of a person, across different datasets. 
As mentioned in Fig.~3 (main paper), $T_z$ has significant impact on the level of perspective distortion observed in an image and becomes more impactful to 3D HMR, the closer the person gets to the camera.
An ideal evaluation dataset for HMR of strongly perspective images should thus contain a large number of samples with persons within close-range to the camera, which we loosely define to be less than 1.5 meter.

\noindent\textbf{HuMMan~\cite{cai2022humman}}: This dataset is captured in a studio environment. A person stands in the middle of a circle of cameras and performs different actions. 
This dataset is useful for performing 3D reconstruction on human subjects due to its multi-view camera setup. 
However, it is very limited in terms of visual diversity due to it being captured in the same studio environment.
More importantly, as shown in Fig.~\ref{fig:distribution_histogram} (red distribution), this dataset contains very limited variation in terms of $T_z$, distributed closely around 1.9m, farther from the close range of $<$1.5m distance.
Therefore, due to its limited visual diversity, $T_z$ variation, and the absence of close-range data with $T_z<1.5$m, this dataset is not ideal for evaluating close-range HMR methods intended to operate on images in-the-wild. Performance on it, thus, is not reflective of performance on highly unconstrained images in the real world.

\noindent\textbf{SPEC-MTP~\cite{kocabas2021spec}}: This dataset is captured using smartphones in the real world with diverse identities, lighting conditions, and poses. 
It is captured by having one person move the camera around another person as they pose for the camera.
3D pose labels are then generated from the video frames. 
As shown in Fig.~\ref{fig:distribution_histogram} (yellow distribution), SPEC-MTP's $T_z$ values fall within the desired 1.5m threshold and center around 1m. 
This $T_z$ distribution and the appearance diversity from real-world capture settings makes SPEC-MTP\cite{kocabas2021spec} a good dataset for evaluating close-range HMR methods.
We find the provided labels to be mostly accurate, while inevitable errors in calibration and video-based reconstruction lead to inaccurate pose labels in a small portion of the test samples.

\noindent\textbf{PDHuman~\cite{wang2023zolly}}: This is a synthetic dataset generated using 630 photogrammetry-scanned human models from Renderpeople~\cite{renderpeople} and animated using Mixamo~\cite{mixamo}. 3D labels are converted to SMPL by optimizing for a set of pose and shape parameters that best fit the 3D human models used to generate the rendered data.
As shown in Fig.~\ref{fig:distribution_histogram} (blue distribution), PDHuman's $T_z$ values are mostly within 1m, leading to high levels of perspective distortion in this dataset.
However, we find that a high percentage of its pose labels are inaccurate with respect to the input images. In Fig.~\ref{fig:bad_gt}, we visualize the SMPL labels overlaid on top of the corresponding images. The SMPL renderings (gray overlays) are generated by using the scripts provided for IoU calculations in the authors' original code base.
We postulate that this inaccuracy may have been the result of inaccurate conversion from the animated RenderPeople models to SMPL.

Considering that quantitative results on PDHuman may not also correctly reflect actual performance, we conclude that there is a lack of accurate and diverse data to quantitatively benchmark performance of close-range HMR for images taken at a $T_z$ depth closer than 1m.
Therefore, we curate a new dataset with accurate labels to facilitate evaluation of close-range HMR.

\subsection{\textsc{Bedlam-cc}: A Close-Range Synthetic Dataset with Accurate 3D Labels}

We create a new close-range evaluation dataset utilizing assets provided with the \textsc{Bedlam} dataset~\cite{black2023bedlam} and name our dataset \textsc{Bedlam-cc}. 
As discussed in the main paper, perspective distortion is non-linear w.r.t. the distance between the camera and the subject~\cite{nagano2019deep}. In particular, it changes rapidly when the distance gets closer (0.3m to 1.2m), because of its inverse relationship to distance.
The perspective projection gradually approximates orthographic projection at distances of 5m and higher.
Therefore, to concentrate our evaluation on close-range HMR, we enforce that $80\%$ of our dataset locates $T_z$ within the range of 0.5m $\leq T_z \leq$ 1.2m and the remaining samples are in the range of 1.2m $< T_z \leq$ 10m.
From the 2 million generated images there are a total of 1314 images in the evaluation split.

We carefully curate the camera poses in our dataset to generate images with diverse viewpoints relative to the person.
With a $T_z$ value being sampled as described above, the camera is positioned on a sphere with the radius given by $T_z$ and randomly sampled spherical coordinates $\theta \in [0,2\pi]$ and $\phi \in [0.1\pi, 0.7\pi]$, where $\theta$ is the azimuth angle and $\phi$ represents the elevation. The camera rotation is evaluated by a LookAt() function towards a randomized target bone along the SMPL-X spine given by a randomized bone index $i \in [0,3,6,9,12,15]$ and an added random noise vector $v \in \mathbb{R}^3$.
To keep the person at a reasonable size relative to the frame we set the focal length using a dolly zoom with a default value $f_d$ of 15mm at 1m distance with a camera sensor size of 36x36mm. We then uniformly randomize the focal length $f_{GT} \in [0.7, 1.3]\cdot f_d$.
In addition, we randomize the lighting setup including skylight (background image and intensity), and directional sun light (position, color, intensity).
We show example images of our \textsc{Bedlam-cc} dataset in Figure~\ref{fig:bedlamcc5x5}. 
Since our dataset is generated through SMPL-X and Unreal Engine, we do not need to convert the data to SMPL-X format and thus avoid conversion errors.

\section{Additional Quantitative Results}
\label{sec:results}

In this section, we report additional quantitative results for various evaluation datasets using more metrics. 
Specifically, we test the various methods on the \textsc{SPEC-MTP}~\cite{kocabas2021spec}, \textsc{PDHuman}~\cite{wang2023zolly}, \textsc{Bedlam-cc}, and \textsc{HuMMan}~\cite{cai2022humman} datasets.
We use the commonly used metrics, including, Mean Per-Joint Position Error (MPJPE), Procrustes Analysis Mean Per-Joint Position Error (PA-MPJPE), Per-Vertex Error (PVE), mean Intersection over Union (mIoU), and Body Part mean Intersection over Union (P-mIoU).
As discussed in the main paper, we introduce new metrics to evaluate the accuracy of recovered perspective projection parameters.
Specifically, we measure the accuracy of the recovered focal length as its percentage error relative to the ground truth focal length:
\begin{equation}
    E_{f} = |f_{pred} - f_{GT}|/f_{GT}.
\end{equation}
Given that $T_z$ has an inverse relationship with respect to the amount of distortion in the image (Fig. 3, main paper), whereas $(T_x,T_y)$ do not, we separately evaluate $T_z$ and $(T_x,T_y)$ errors as $E_{T_z}$ and $E_{T_{xy}}$ in meters. 
Additionally, since $T_z$'s accuracy is less important at far distances, we also calculate an inverse $T_z$ error $E_{1/T_z}$, reflecting this property:
\begin{align}
    E_{T_{xy}} &= \|T_{xy}^{pred} - T_{xy}^{GT}\|_2,\\
    E_{T_z} &= |T_z^{pred} - T_z^{GT}|, \\
    E_{1/T_z} &= |1/T_z^{pred} - 1/T_z^{GT}|.
\end{align}

\begin{figure*}[t!]
    \centering
    \includegraphics[width=0.79\textwidth, trim=0 120 1400 0, clip]{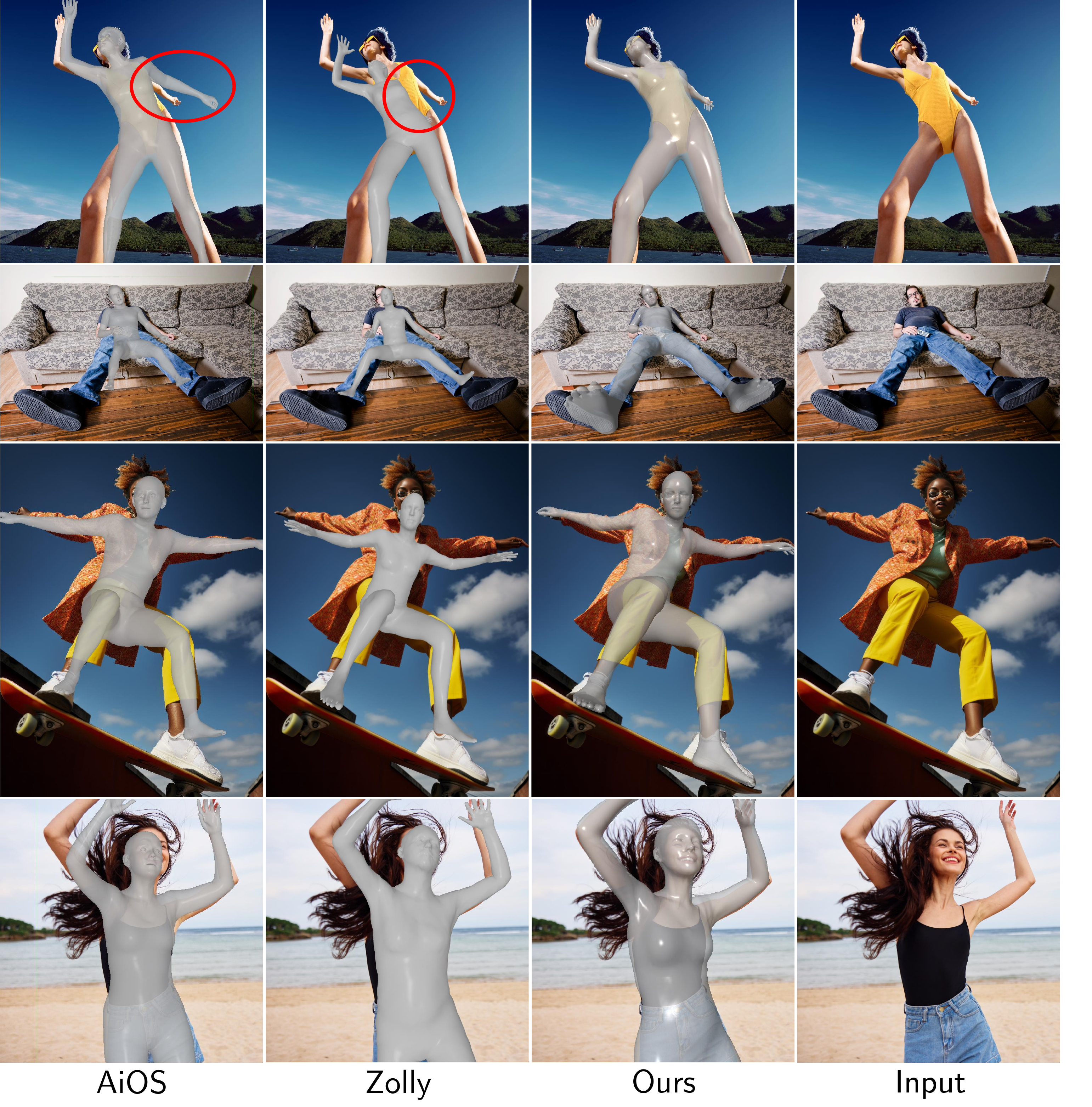}
    \caption{\textbf{More Qualitative Results.} Beyond accurate pose estimation, our approach BLADE effectively reconstructs perspective projection parameters, allowing the predicted 3D human mesh to align closely with the input image.}
    \label{fig:vis_3}
\end{figure*}

\begin{figure*}[t!]
    \centering
    \includegraphics[width=0.8\textwidth, trim=0 330 1400 0, clip]{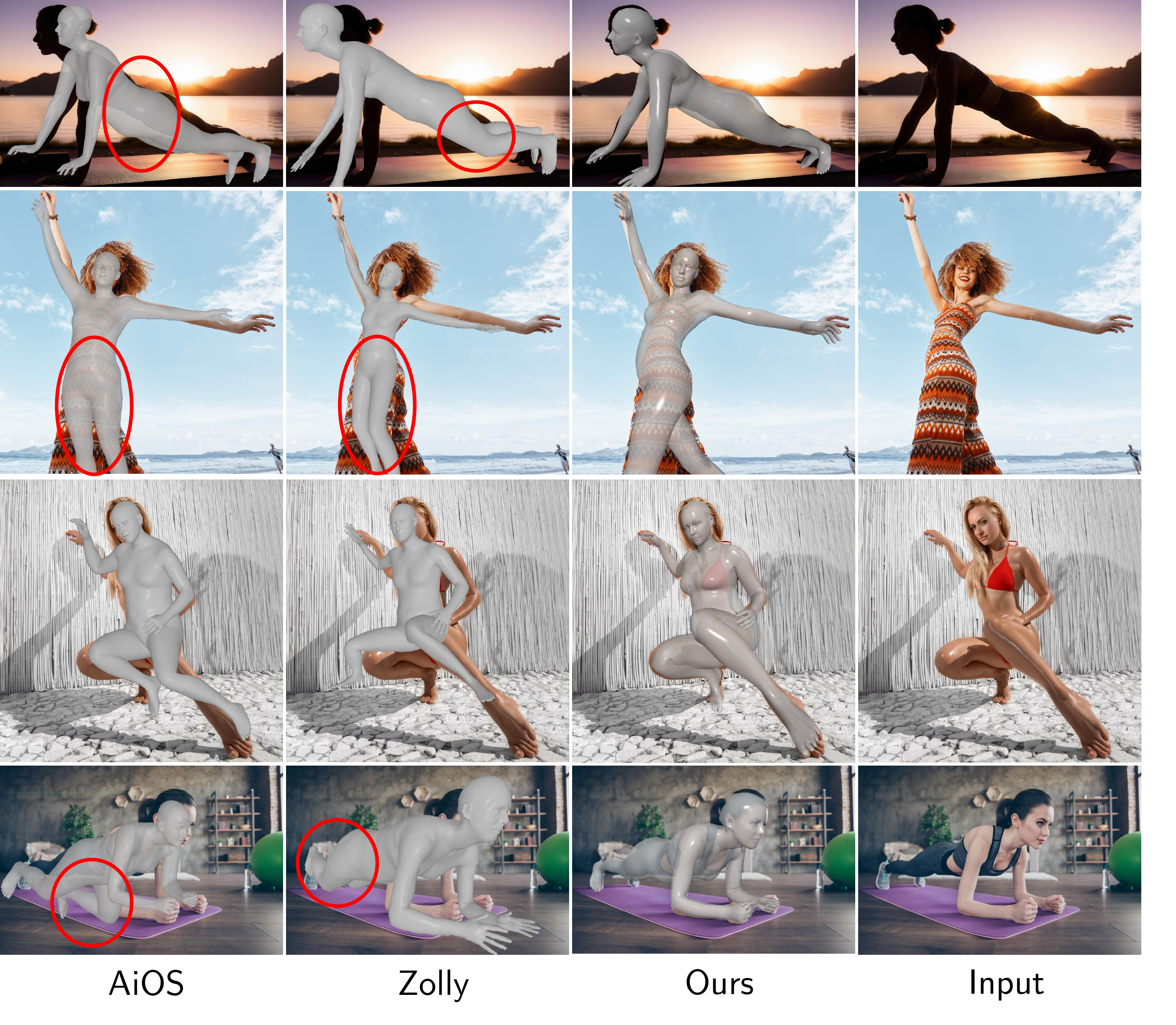}
    \caption{\textbf{More Qualitative Results.} Our approach BLADE not only estimates 3D shape and pose precisely but also accurately retrieves perspective projection parameters, enabling the predicted 3D human mesh to align seamlessly with the input image.}
    \vspace{-1em}
    \label{fig:vis_4}
\end{figure*}

\begin{figure*}[t!]
    \centering
    \includegraphics[width=0.8\textwidth, trim=0 300 1400 0, clip]{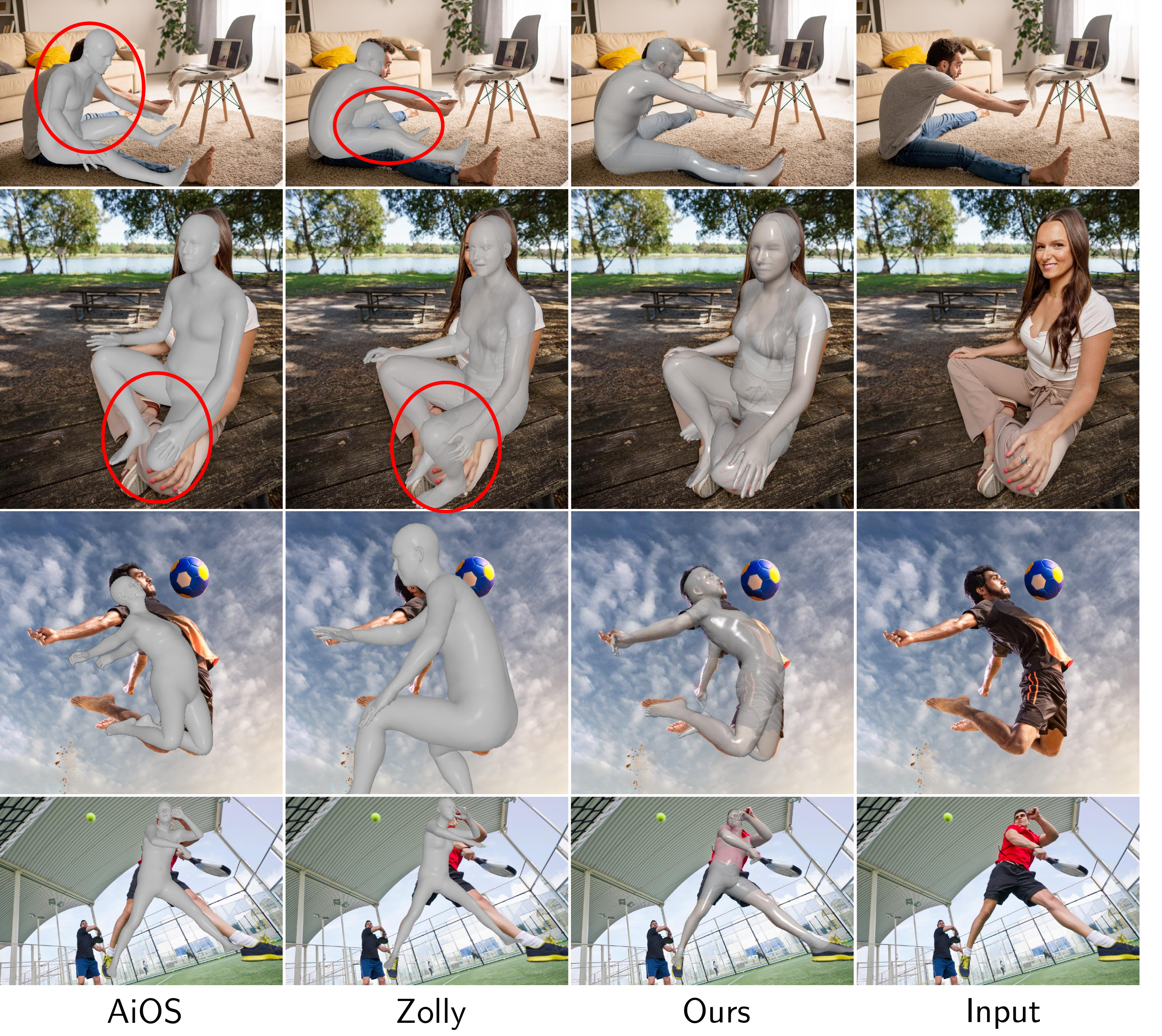}
    \caption{\textbf{More Qualitative Results.} BLADE not only achieves accurate pose estimation, but also recovers accurate perspective projection parameters and thus can align the predicted 3D human mesh to the input image well.}
    \label{fig:vis_5}
\end{figure*}

\begin{table*}[ht]
    \scalebox{0.68}{\input{tables_ours/results_supplemental}}
    \caption{\textbf{Results of SOTA methods} on the \textsc{SPEC-MTP}~\cite{kocabas2021spec}, \textsc{PDHuman}~\cite{wang2023zolly}, \textsc{Bedlam-CC}, and \textsc{HuMMan}~\cite{cai2022humman} datasets. 
    For baselines at the top of the tables, we use the results reported by Zolly~\cite{wang2023zolly} and omit the ones not available.
    Additionally, we re-evaluate newer state-of-the-art methods AiOS~\cite{sun2024aios}, SMPLer-X~\cite{cai2023smplerx}, and TokenHMR~\cite{dwivedi_cvpr2024_tokenhmr}.
    These models are noted using "*".
    }
    \label{tab:sota}
\end{table*}

In Table.~\ref{tab:sota}, we show that BLADE achieves state-of-the-art accuracy for a majority of the metrics across the four datasets: \textsc{SPEC-MTP}\cite{kocabas2021spec}, \textsc{PDHuman}\cite{wang2023zolly}, \textsc{Bedlam-cc}, and \textsc{HuMMan}\cite{cai2022humman}.
Among these \textsc{SPEC-MTP}\cite{kocabas2021spec}, \textsc{PDHuman}\cite{wang2023zolly}, and \textsc{Bedlam-cc} are perspectively distorted datasets with many persons with $T_z<1.5$m.
On perspectively distorted datasets, BLADE is state of the art in terms of recovering accurate perspective projection parameters (measured by $E_{T_z}$, $E_{1/T_z}$, $E_{T_xy}$, and $E_{f}$) and accurate 3D mesh recovery (measured by PVE).
Additionally, BLADE achieves joint accuracies (measured by PA-MPJPE and MPJPE) better than or comparable to state-of-the-art methods.
The accurate recovery of projection parameters and 3D geometry results in state-of-the-art alignment from the rendered mesh to the input image.
This is shown by BLADE's significantly higher mIoU and P-mIoU performances.
For example, on \textsc{SPEC-MTP}\cite{kocabas2021spec}, BLADE's mIoU is $69.9\%$, whereas the second best method PARE\cite{kocabas2021pare} achieves $55.8\%$. 
Similarly, on \textsc{PDHuman}~\cite{wang2023zolly} and \textsc{Bedlam-cc}, BLADE achieves mIoU values of $67.3\%$ and $72.8\%$, respectively, whereas the second best methods achieve $53.0\%$ and $54.6\%$.
Moreover, BLADE consistently achieves high IoU values of around $70\%$, whereas prior methods show significant degradation on the three perspectively distorted datasets.
On the less distorted \textsc{HuMMan}\cite{cai2022humman} dataset, we achieve state-of-the-art accuracy on $T_z$ estimation ($E_{T_z}$, $E_{1/T_z}$) and focal length estimation ($E_f$). 
BLADE achieves significantly better joint precisions (PA-MPJPE, MPJPE) and 3D mesh reconstruction than the recent state-of-the-art methods (AiOS\cite{sun2024aios}, SMPLer-X\cite{cai2023smplerx}, and TokenHMR\cite{dwivedi_cvpr2024_tokenhmr}) and is comparable to Zolly.

\section{Single-Image Ambiguity in 3D Human Mesh Recovery (3D HMR)}
\begin{figure}[t!]
    \centering
    \includegraphics[width=0.5\textwidth, trim=0 0 0 0, clip]{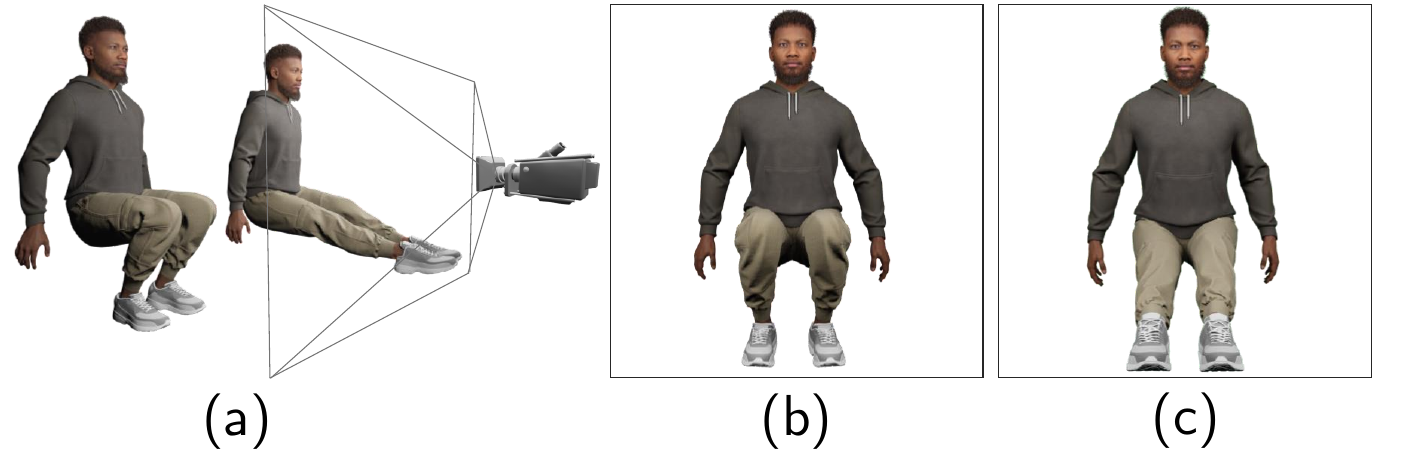}
    \caption{\textbf{The Ambiguity of Single Image 3D Human Pose Estimation.} Although being significantly different in pose and distance to the camera (a) both presented configurations result in similar camera views (b, c). Therefore, due to the ill-posed nature of single-image 3D pose estimation, different combinations of pose and camera distance can result in valid but incorrect reconstructions.}
    \label{fig:poseambiguity}
\end{figure}

\begin{figure}[t!]
    \centering
    \includegraphics[width=0.48\textwidth, trim=0 0 0 0, clip]{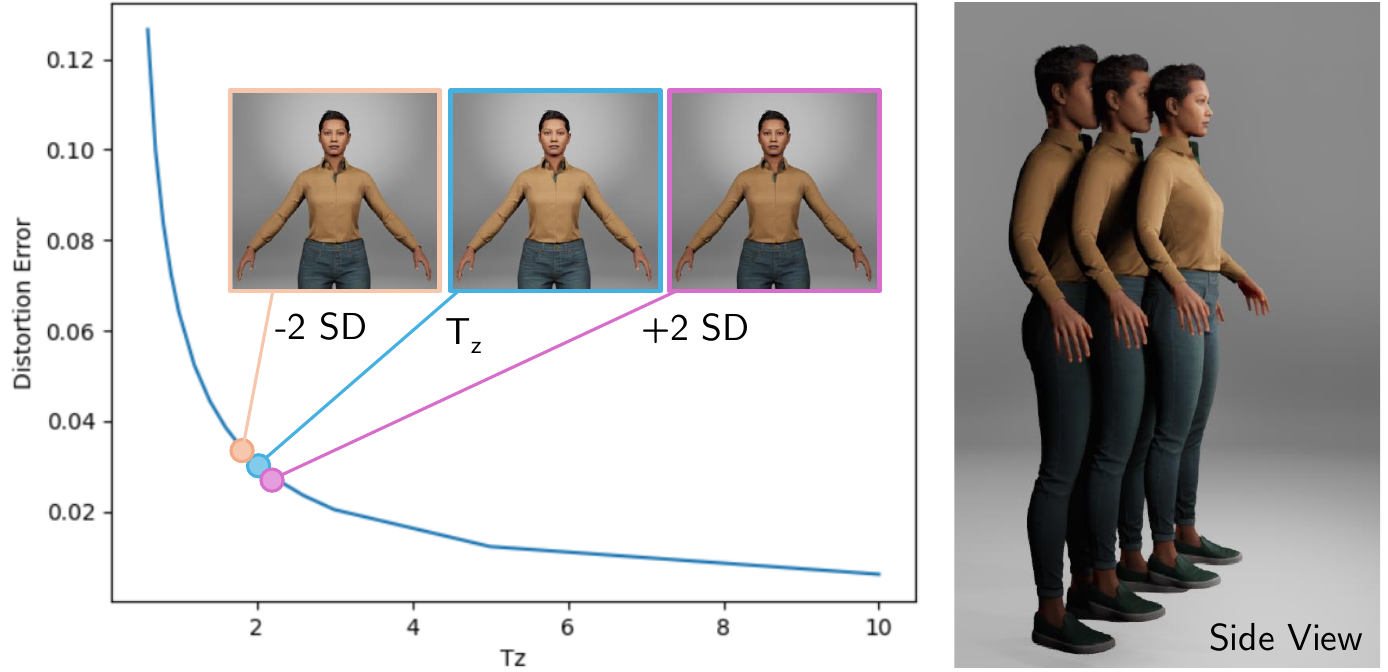}
    \caption{\textbf{Ambiguous Human Size from a Single Image.} 
    The problem of metric-scale mesh estimation problem is inherently ill-posed, and capturing people of different sizes from different distances can result in similar images. 
    The side view reveals the actual sizes of the subjects and their distances $T_z$ to the camera. 
    When the image of a taller person captured farther away can be similar to the image of a shorter person captured at a closer distance.
    The corresponding $T_z$ values are also shown on the left.
    However, given that the heights of 95 \% of all human~\cite{fryar2012anthropometric} ($\pm2$  standard deviations) lie within a small range, the size variation thus correspond to a narrow $T_z$ variation as shown on the left curve.
    The mean size is the blue inset and the range of $\pm2$  standard deviations are shown as yellow and violet insets. 
    }
    \label{fig:scaleambiguity}
\end{figure}

\begin{figure*}[ht]
  \centering
\includegraphics[width=0.97\linewidth, trim=0pt 180pt 0pt 0pt, clip]{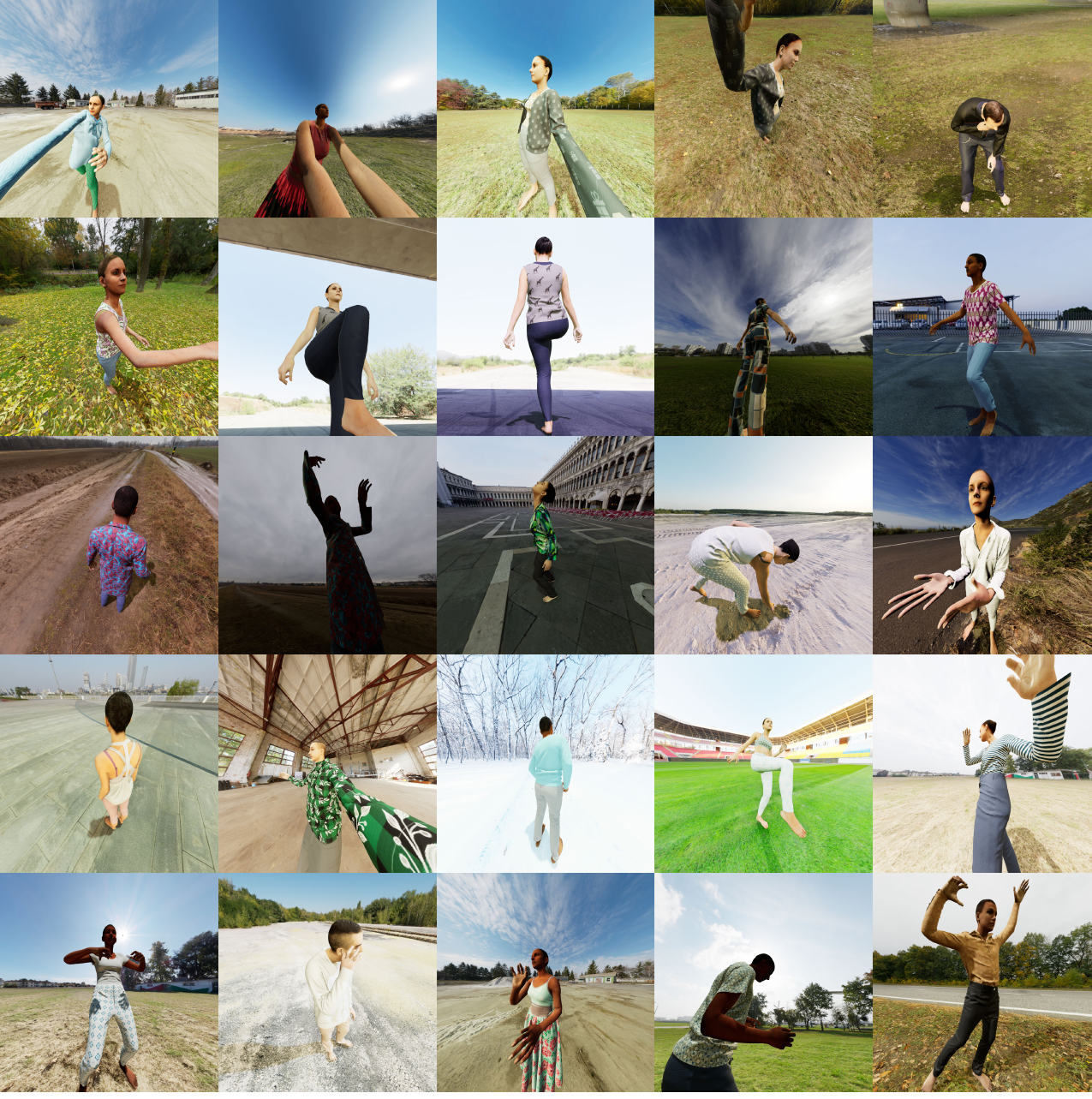}
  \caption{\textbf{Examples of our synthetic \textsc{Bedlam-cc} dataset.} The strong variation in lighting and camera angles as well as occasional extreme close-up distortion are intentionally part of the data.}
  \label{fig:bedlamcc5x5}
\end{figure*}

In Fig.~\ref{fig:poseambiguity} and \ref{fig:scaleambiguity}, we visually illustrate the ambiguity in single-image human mesh recovery.
To achieve both accurate 3D mesh recovery and 2D alignment, one needs to solve for both the 3D mesh of the person as well as the camera intrinsic and extrinsic parameters.
However, given that none of the aforementioned parameters is known, and that they are heavily entangled, this problem is well known to be ill-posed and has potentially infinite solutions.
For example, as shown in Fig.~\ref{fig:poseambiguity}, it is difficult for a model to correctly predict the two poses from the input images because it has no information about the shape of the person's legs and shoes.
Moreover, due to the nature of projected geometry, the reconstructions are always up to scale unless additional knowledge of scale is provided, \eg the camera's movement is measured in physical units.
For example, as shown in Fig.~\ref{fig:scaleambiguity}, images of people of different sizes can result in very similar images. Therefore, the reverse problem of reconstructing the person from the images can also result in 3D meshes of different sizes.

While the aforementioned ambiguities are inherent to the problem, much prior work have leveraged the regularity of the human body to arrive at reasonable solutions for this ill-posed problem.
For example, one such regularity~\cite{owid-human-height} is that $95\%$ of men have a height between 163.2cm and 193.6cm and $95\%$ of women have a height between 150.6cm and 178.84cm.

\label{sec:ambiguity}

\section{Trade-Off between Close and Far Range $T_z$ Estimation}
\label{sec:depthtradeoff}
For $T_z$ estimators trained without our \textsc{Bedlam-cc} dataset, we observe that it is difficult for them to achieve accurate $T_z$ estimation for both close and far range images.
The various $T_z$ estimators with different backbones oscillate between achieving high accuracy on close-range or on far-range images, exemplified by their accuracies on the close range dataset \textsc{SPEC-MTP}~\cite{kocabas2021spec} and the farther range dataset \textsc{HuMMan}~\cite{cai2022humman}.
For example, when using Sapiens\cite{khirodkar2025sapiens} as the backbone for our $T_z$ estimator, its best $T_z$ error on \textsc{SPEC-MTP}\cite{kocabas2021spec} is 21cm, but it scores a high $T_z$ error of 70cm on\textsc{HuMMan}.
On the other hand, using a model checkpoint with a low $T_z$ error of 60cm on \textsc{HuMMan} results in an 85cm error on \textsc{SPEC-MTP}.
Similarly, when using DepthAnythingV2~\cite{depth_anything_v2} as the backbone, our $T_z$ estimator can achieve a low $T_z$ error of 15.4cm on \textsc{SPEC-MTP}~\cite{kocabas2021spec}, but at the same time suffers from a high $T_z$ error of 23cm on \textsc{HuMMan}~\cite{cai2022humman}.
When using a checkpoint that can achieve 3.1cm $T_z$ error on \textsc{HuMMan}, the model in turn suffers from a high $T_z$ error of 67.6 on \textsc{SPEC-MTP}.

Inspired by recent works in monocular depth estimation~\cite{metric3d,depth_anything_v2}, we focus on providing the networks with more high quality close-range training samples by curating our own \textsc{Bedlam-cc} dataset (Sec.~\ref{sec:datasetsextended}).
With more high quality close-range training samples, our final $T_z$ estimator achieves a low error of 12.7cm on the close-range dataset \textsc{SPEC-MTP}~\cite{kocabas2021spec} while maintaining a reasonable $T_z$ error of 18.7cm on the farther-range \textsc{HuMMan} dataset (Table.~\ref{tab:sota}).

\paragraph{Dataset license information.}
The assets of the \textsc{Bedlam} dataset~\cite{black2023bedlam} have been published by Max Planck Institute for Intelligent Systems under a \textit{No distribution} license\footnote{https://bedlam.is.tuebingen.mpg.de/license.html}.

With the publication of our work we will publish
\begin{itemize}
    \item our code changes with respect to the \textsc{Bedlam} dataset to render the \textsc{Bedlam-cc} dataset, and
    \item instructions to render the \textsc{Bedlam-cc} dataset.
\end{itemize}
For recreation of the \textsc{Bedlam-cc} dataset the render pipeline needs to be setup according to the guidelines of the \textsc{Bedlam} dataset.
We will publish our data under license terms to allow usage for research purposes.

~\newpage

\section*{Image Sources}
\begin{itemize}
\item Main Paper Figure 1: Adobe Stock image ids: 16532441, 688449553, 868801378.\footnote{https://stock.adobe.com/}
\item Main Paper Figure 4: Adobe Stock Image id: 789510049.
\item Main Paper Table 1: Row 1-2 Adobe Stock image ids: 415527042, 344928073, 71230339, 605587274. Last row: Images from Zolly~\cite{wang2023zolly}.
\item Figure A1: Adobe Stock image ids: 184701266, 21677394, 60240732.
\item Figure A4: Adobe Stock image ids: 859644245, 81892568, 21197764, 
902825438.
\item Figure A5: Adobe Stock image ids: 892029686, 71230339, 688449514, 615119495.
\item Figure A6: Adobe Stock image ids: 1061297360, 765162341, 547882981, 355426702.
\item Figure A7: Adobe Stock image ids: 348174880, 583910785, 219801712, 63038620.
\end{itemize}

%% file: tables_ours/results_supplemental.tex
%
\begin{tabular}{lccccccccc|ccccccccc}
\toprule
\multirow{2}{*}{\textbf{Methods}} & \multicolumn{9}{c}{\textsc{SPEC-MTP}~\cite{kocabas2021spec} (real-world capture)}& \multicolumn{9}{c}{\textsc{PDHuman}~\cite{wang2023zolly} (synthetic)}\\ \cmidrule{2-19}
&$E_{T_z}$↓&$E_{1/T_z}$↓&$E_{T_{xy}}$↓&$E_{f}$↓&PA-MPJPE$\downarrow$&MPJPE$\downarrow$& PVE$\downarrow$&mIoU$\uparrow$&P-mIoU$\uparrow$&$E_{T_z}$↓&$E_{1/T_z}$↓&$E_{T_{xy}}$↓&$E_{f}$↓&PA-MPJPE$\downarrow$ & MPJPE$\downarrow$&PVE$\downarrow$& mIoU$\uparrow$&P-mIoU$\uparrow$\\ \midrule 

\rule{0pt}{10pt} HMR~\cite{kanazawaHMR18}                       &-&-&-&-            & 73.9 & 121.4 & 145.6 & 48.8 & 16.0   &-&-&-&-             & 62.5 & 91.5 & 106.7 & 48.9 & 21.7 \\

\rule{0pt}{10pt} HMR-$f$~\cite{kanazawaHMR18}                   &-&-&-&-            & 72.7 & 123.2 & 145.1 & 52.3 & 20.1   &-&-&-&-             & 61.6 & 90.2 & 105.5 & 45.2 & 20.4   \\

\rule{0pt}{10pt} SPEC~\cite{kocabas2021spec}                    &-&-&-&-            & 76.0 & 125.5 & 144.6 & 49.9 & 18.8   &-&-&-&-             & 65.8 & 94.9 & 109.6 & 43.4 & 19.6  \\

\rule{0pt}{10pt} CLIFF~\cite{li2022cliff}                       &-&-&-&-            & 74.3 & 115.0 & 132.4 & 53.6 & 23.7   &-&-&-&-             & 66.2 & 99.2 & 115.2 & 51.4 & 24.8 \\

\rule{0pt}{10pt} PARE~\cite{kocabas2021pare}                    &-&-&-&-            & 74.2 & 121.6 & 143.6 & 55.8 & 23.2   &-&-&-&-             & 66.3 & 95.9 & 116.7 & 48.2 & 20.9 \\

\rule{0pt}{10pt} GraphCMR~\cite{kolotouros2019convolutional}    &-&-&-&-            & 76.1 & 121.4 & 141.6 & 53.5 & 22.0   &-&-&-&-             & 62.0 & 85.8 & 98.4  & 47.9 & 21.5  \\

\rule{0pt}{10pt} FastMETRO~\cite{cho2022cross}                  &-&-&-&-            & 75.0 & 123.1 & 137.0 & 53.5 & 20.5   &-&-&-&-             & 58.6 & 83.6 & 95.4  & 50.1 & 22.5 \\

\rule{0pt}{10pt} Zolly~\cite{wang2023zolly}         & 0.899 & 0.394 & 0.906 & 106.3    & 67.4 & 114.6 & 126.7 & 62.3 & 30.4    &0.255 & 0.355 & 0.051 & 27.3    & 49.9 & 70.7 & 82.0  & 53.0 & 26.5 \\

\rule{0pt}{10pt} SMPLer-X*                         &0.980 & 0.450 & 0.109 & 112.1    & \textbf{55.5} & \textbf{90.9} & 102.6 & 53.0 & 15.9     &2.223 & 1.030 & 0.126 & 55.0    & 96.8 & 148.2 & 161.2 & 47.6 & 17.1 \\

\rule{0pt}{10pt} TokenHMR*                         & 0.909 & 0.436 & 0.095 & 112.1    & 64.2 & 107.1 & 124.3 & 49.8 & 19.0     &2.280 & 1.034 & 0.068 & 55.0     & 92.1 & 141.5 & 156.7 & 53.0 & 27.8 \\

\rule{0pt}{10pt} AiOS*                            & 1.035 & 0.464 & 0.121 & 112.1    & 62.8 & 101.6 & 110.9 & 48.7 & 11.3     &2.312 & 1.024 & 0.149 & 55.0     & 106.6 & 170.6 & 183.4 & 49.5 & 16.0 \\

\hline
\rule{0pt}{10pt} Ours                             & 0.129 & 0.114 & 0.056 & 16.3    & 61.0 & 105.3 & 111.9 & 68.6 & 39.8     &\textbf{0.106} & \textbf{0.176} & \textbf{0.043} & \textbf{21.6}     & \textbf{49.6} & \textbf{69.7} & \textbf{80.5} & \textbf{67.3} & \textbf{44.6} \\

\rule{0pt}{10pt} Ours (real-world)                & \textbf{0.127} & \textbf{0.112} & \textbf{0.044} & \textbf{15.9}    & 56.7 & 94.1 & \textbf{99.6} & \textbf{69.9} & \textbf{41.5}     &0.107 & 0.178 & 0.049 & 22.3     & 61.4 & 90.1 & 102.6 & 65.2 & 41.4 \\ \midrule

\multirow{2}{*}{} & \multicolumn{9}{c}{\textsc{Bedlam-cc} (synthetic)}& \multicolumn{9}{c}{\textsc{HuMMan}~\cite{cai2022humman} (studio capture)}\\ \cmidrule{2-19}
&$E_{T_z}$↓&$E_{1/T_z}$↓&$E_{T_{xy}}$↓&$E_{f}$↓&PA-MPJPE$\downarrow$&MPJPE$\downarrow$& PVE$\downarrow$&mIoU$\uparrow$&P-mIoU$\uparrow$&$E_{T_z}$↓&$E_{1/T_z}$↓&$E_{T_{xy}}$↓&$E_{f}$↓&PA-MPJPE$\downarrow$ & MPJPE$\downarrow$&PVE$\downarrow$& mIoU$\uparrow$&P-mIoU$\uparrow$\\ 

\rule{0pt}{10pt} HMR~\cite{kanazawaHMR18}                       &-&-&-&-            &-&-&-&-&-    &-&-&-&-            & 30.2 & 43.6 & 52.6 & 65.1 & 39.5 \\

\rule{0pt}{10pt} HMR-$f$~\cite{kanazawaHMR18}                   &-&-&-&-            &-&-&-&-&-    &-&-&-&-            & 29.9 & 43.6 & 53.4 & 62.7 & 34.9  \\

\rule{0pt}{10pt} SPEC~\cite{kocabas2021spec}                    &-&-&-&-            &-&-&-&-&-    &-&-&-&-            & 31.4 & 44.0 & 54.2 & 51.4 & 25.6 \\

\rule{0pt}{10pt} CLIFF~\cite{li2022cliff}                       &-&-&-&-            &-&-&-&-&-    &-&-&-&-            & 28.6 & 42.4 & 50.2 & 68.8  & 44.7\\

\rule{0pt}{10pt} PARE~\cite{kocabas2021pare}                    &-&-&-&-            &-&-&-&-&-    &-&-&-&-            & 32.6 & 53.2 & 65.5 & 66.5 & 38.3 \\

\rule{0pt}{10pt} GraphCMR~\cite{kolotouros2019convolutional}    &-&-&-&-            &-&-&-&-&-    &-&-&-&-            & 29.5 & 40.6 & 48.4 & 61.6 & 37.5 \\

\rule{0pt}{10pt} FastMETRO~\cite{cho2022cross}                  &-&-&-&-            &-&-&-&-&-    &-&-&-&-            & 26.3 & 38.8 & 45.5 & 68.3 & \textbf{45.2} \\

\rule{0pt}{10pt} Zolly~\cite{wang2023zolly}        &0.539 & 0.634 & 0.081 & 46.1    & 68.8 & 107.8 & 131.8 & 51.8 & 21.2   &0.228 & 0.072 & 0.034 & 9.4    & \textbf{22.3} & \textbf{32.6} & \textbf{40.0} & 71.2 & 45.1\\

\rule{0pt}{10pt} SMPLer-X*                         &2.057 & 1.172 & 0.087 & 134.9    & 69.5 & 120.3 & 140.0 & 53.0 & 21.3    &2.461 & 0.300 & 0.125 & 41.6    & 38.7 & 56.4 & 65.8 & 51.8 & 11.1 \\

\rule{0pt}{10pt} TokenHMR*                         &2.378 & 1.200 & 0.096 & 134.9    & 59.9 & 114.3 & 136.4 & 54.1 & 22.3    &2.599 & 0.307 & 0.044 & 41.6    & 46.4 & 72.2 & 82.0 & 60.9 & 31.1\\

\rule{0pt}{10pt} AiOS*                            &2.340 & 1.197 & 0.111 & 134.9    & 71.6 & 125.7 & 143.0 & 54.6 & 19.9    &2.311 & 0.292 & \textbf{0.033} & 41.6     & 66.1 & 91.8 & 99.4 & \textbf{72.0} & 44.3\\

\hline
\rule{0pt}{10pt} Ours                           &0.326 & 0.306 & 0.066 & 26.2    & 59.4 & 90.5 & 111.6 & 72.7 & \textbf{44.5}    & 0.188 & \textbf{0.058} & 0.055 & 8.5    & 24.9 & 44.4 & 56.3 & 69.8 & 37.9 \\

\rule{0pt}{10pt} Ours (real-world)              &\textbf{0.325} & \textbf{0.305} & \textbf{0.065} & \textbf{25.7}    & \textbf{57.8} & \textbf{85.8} & \textbf{106.8} & \textbf{72.8} & \textbf{44.5}    &\textbf{0.187} &\textbf{0.058} & 0.056 & \textbf{8.3}    & 23.8 & 41.1 & 52.3  & 70.6 & 38.2 \\ 

\bottomrule

\end{tabular}